\documentclass[letterpaper]{article} 
\usepackage[submission]{aaai25}  
\usepackage{times}  
\usepackage{helvet}  
\usepackage{courier}  
\usepackage[hyphens]{url}  
\usepackage{graphicx} 
\usepackage{natbib}  
\usepackage{caption} 
\usepackage{amsmath} 
\usepackage{amssymb}
\usepackage[most]{tcolorbox}
\usepackage{colortbl}
\usepackage{soul}     
\usepackage{xcolor}   
\definecolor{grey}{gray}{0.85}
\usepackage{algorithm}
\usepackage{algorithmic}
\usepackage{multirow}
\usepackage{multicol}
\usepackage{color}
\definecolor{lightred}{rgb}{1, 0.8, 0.8}
\definecolor{lightgreen}{rgb}{0.8, 1, 0.8}
\definecolor{lightblue}{rgb}{0.8, 0.9, 1}
\usepackage{longtable} 

\newcommand*\colourcheck[1]{%
  \expandafter\newcommand\csname #1check\endcsname{\textcolor{#1}{\ding{52}}}%
}
\colourcheck{blue}
\colourcheck{green}
\colourcheck{red}
\usepackage{pifont}

\usepackage{pdfpages}

\usepackage{newfloat}
\usepackage{listings}
\DeclareCaptionStyle{ruled}{labelfont=normalfont,labelsep=colon,strut=off} 
\floatstyle{ruled}
\newfloat{listing}{tb}{lst}{}
\floatname{listing}{Listing}
\title{IdeaBench: Benchmarking Large Language Models for Research Idea Generation}
\author{
    Sikun Guo\equalcontrib, Amir Hassan Shariatmadari\equalcontrib, Guangzhi Xiong, Albert Huang, Eric Xie, Stefan Bekiranov, Aidong Zhang
}
\affiliations{
    University of Virginia\\

    \{qkm6sq, ahs5ce, hhu4zu, kfa7fg, jrg4wx, sb3de, aidong\}@virginia.edu
}

\makeatletter
\def\showauthors@on{T}
\makeatother

\begin{document}

\maketitle

\begin{abstract}

Large Language Models (LLMs) have transformed how people interact with artificial intelligence (AI) systems, achieving state-of-the-art results in various tasks, including scientific discovery and hypothesis generation. However, the lack of a comprehensive and systematic evaluation framework for generating research ideas using LLMs poses a significant obstacle to understanding and assessing their generative capabilities in scientific discovery. To address this gap, we propose \textit{IdeaBench}, a benchmark system that includes a comprehensive dataset and an evaluation framework for standardizing the assessment of research idea generation using LLMs. Our dataset comprises titles and abstracts from a diverse range of influential papers, along with their referenced works. To emulate the human process of generating research ideas, we profile LLMs as domain-specific researchers and ground them in the same context considered by human researchers. This maximizes the utilization of the LLMs' parametric knowledge to dynamically generate new research ideas. We also introduce an evaluation framework for assessing the quality of generated research ideas. Our evaluation framework is a two-stage process: first, using GPT-4o to rank ideas based on user-specified quality indicators such as novelty and feasibility, enabling scalable personalization; and second, calculating relative ranking based ``Insight Score" to quantify the chosen quality indicator. The proposed benchmark system will be a valuable asset for the community to measure and compare different LLMs, ultimately advancing the automation of the scientific discovery process. Our code and dataset are available at: \url{https://anonymous.4open.science/r/IdeaBench-2747/}.
\end{abstract}

\section{Introduction}

Recent years have witnessed the rapid development of Large Language Models (LLMs). LLMs like GPT-4 \cite{OpenAI_GPT4_2023} and LLama series \cite{touvron2023llama} introduced advanced capabilities that set them apart from previous generations of machine learning models. Among these capabilities, in-context learning allows LLMs to understand and respond to user prompts in a nuanced manner without requiring additional training for each specific task, enabling LLMs to generalize across a wide range of tasks, providing robust state-of-the-art performance even with limited data \cite{brown2020language}. As a result, LLMs have revolutionized the way humans interact with AI systems, making it possible to generate coherent text, translate languages, answer questions, and even compose creative content with unprecedented accuracy and fluency \cite{bubeck2023sparks}. The impact of these advancements extends beyond consumer applications, influencing various sophisticated domains such as education \cite{moore2023empowering}, healthcare \cite{yang2023large}, and scientific research \cite{wysocki2024llm}. 

Recently, the impressive performance of LLMs in everyday applications has sparked significant interest in academia, particularly for their potential use in scientific discovery or hypothesis generation \cite{ai4science2023impact}. Several studies have explored leveraging LLMs to generate hypotheses or research ideas \cite{yang2023large2, wang2023scimon, zhou2024hypothesis, baek2024researchagent, qiu2023phenomenal}. However, despite numerous results, a unified and comprehensive framework for evaluating generated research ideas is still lacking, making it difficult for the community to clearly understand the performance spectrum of different techniques for generating research ideas.

To address this limitation, we introduce a standardized evaluation framework designed to emulate how human researchers generate research ideas. This framework, termed {\it IdeaBench}, comprises three main components: dataset construction, research idea generation, and a novel metric to evaluate the quality of the generated research ideas. The intuition behind this framework is grounded in the typical research process of how researchers generate new scientific research ideas as described below:

\begin{enumerate}
    \item Targeting a specific topic.
    \item Reviewing related literature, focusing on recent findings and methodologies.
    \item Identifying gaps in knowledge or methods within these recent findings.
    \item Proposing research ideas to address these gaps.
\end{enumerate}

We first construct a benchmark dataset that includes meticulously filtered 2,374 target papers' 
abstracts 
from biomedical research fields. These target papers serve as the ground-truth sources of research ideas. Additionally, the dataset contains the 
abstracts of the papers referenced by the target papers, providing the context necessary for LLMs to generate relevant research ideas. This comprehensive dataset aims to capture the complexity and specificity of scientific research, particularly in the biomedical domain, thus offering a solid foundation for evaluating LLMs' capability in generating research ideas. 

Based on the benchmark dataset, we design a prompt template that leverages LLMs to generate research ideas. In addition to grounding the context for idea generation using reference papers from our dataset, we also profile the LLMs as domain-specific researchers in the prompt. This approach aims to dynamically maximize the utilization of the LLMs' parametric knowledge, enabling the generation of more in-depth and insightful research ideas. It can also be used as a baseline for future comparisons. 

To accurately assess the quality of generated research ideas, we design an evaluation framework which incorporates two critical components: personalized quality ranking and relative quality scoring. This dual approach allows for a nuanced assessment that takes into account user-defined quality indicators such as novelty, feasibility, etc. Our design ensures a versatile and comprehensive evaluation framework, capable of adapting to different research contexts and providing meaningful insights into the quality of LLM-generated ideas.
Our results show that recent high-capacity LLMs are capable of generating research ideas using \textit{IdeaBench} dataset, and our metric is able to assess the quality of generated research ideas from different dimensions. We hope this work inspires academia to further unleash the potential of LLMs in supporting research ideation, ultimately accelerating 
scientific discovery in the future.

To summarize, our contributions are as follows:
\begin{itemize}

\item We construct \textit{IdeaBench} dataset, which consists of 2,374 influential biomedical target papers along with their 29,408 reference papers, to evaluate LLMs' capabilities in generating research ideas.

\item We propose an evaluation framework which offers a scalable and versatile metric called ``Insight Score", which can quantify novelty, feasibility, or any other quality indicators defined by human researchers.

\item We conduct extensive experiments to demonstrate several LLMs' abilities in generating research ideas based on our dataset and evaluation metric.

\end{itemize}


\section{Related work}
\subsubsection{Machine Learning for Hypothesis Generation.}
Most existing research on hypothesis generation has concentrated on literature-based discovery (LBD), aiming to predict pairwise relationships between discrete concepts \cite{wang2023learning}. This approach involves uncovering new scientific knowledge by mining literature to identify meaningful implicit associations between unrelated biomedical concepts. The majority of prior studies have focused on identifying these implicit connections from snapshots of the corpus. While these LBD-based approaches are accurate and verifiable, they assume that all concepts are known beforehand and need only to be connected, without considering the contextual factors that human scientists incorporate during ideation. Moreover, these methods do not address the inductive and generative nature of scientific inquiry. Recently, several new studies have explored the use of large language models (LLMs) for hypothesis generation. For instance, in \cite{wang2023scimon}, the authors presented a framework called SciMON that leverages past scientific literature as context for fine-tuning LLMs for hypothesis generation. MOOSE \cite{yang2023large2} utilized multi-level LLM self-feedback to boost scientific hypotheses discovery in social science. ResearchAgent \cite{baek2024researchagent} employed LLMs to automatically generate and refine problems, methods, and experiment designs starting with a core paper and entity-centric knowledge graphs. \cite{zhou2024hypothesis} proposed a prompting approach to iteratively generate hypotheses using LLMs based on training examples.

\subsubsection{Evaluation for Open-ended Text Generation.}
Although human judgment is still considered the golden standard for evaluating open-ended text generation, the Natural Language Processing community has tried to develop different approaches to approximate human evaluation in a scalable way. Traditional metrics like BLEU \cite{papineni2002bleu} and ROUGE \cite{Lin2004ROUGEAP} measure the lexical overlap between model generated content and ground-truth reference. Later on, several efforts use pre-trained language models to measure distributional similarity \cite{zhang2019bertscore, zhao2019moverscore} or token probabilities \cite{yuan2021bartscore, thompson2020automatic}. With the increasing popularity and impressive performance of Large Language Models, recent endeavors employ LLMs as autoraters for open-ended text generation \cite{chiang2023can, liu2023g, bubeck2023sparks, bai2024benchmarking, fu-etal-2024-gptscore, vu2024foundational}, the effectiveness of using LLMs as autoraters is often reflected by its correlation with human-ratings, making autoraters a promising alternative to human evaluators for large-scale evaluation.

\section{Methodology}
In this section, we introduce the details of the three components of our framework, namely, dataset construction, research idea generation, and evaluation of the generated ideas. The first component is to collect a set of valuable target papers and reference papers so that the reference papers can be used to generate new research ideas and compare with those in the target papers. The second component is to design an LLM prompt tailored for generating research ideas,
and the last component is to formulate an evaluation metric to measure the quality of the generated ideas. 

\vspace{-0.08in}
\subsection{Dataset Construction}\label{sec:dataset-construction}
The dataset construction consists of two components: curating a set of valuable papers which will be used as the target papers, and accumulating the reference papers which were used to generate ideas in the target papers.

\subsubsection{Data Collection.}
To create a benchmark dataset for evaluating the research idea generation capabilities of LLMs, we meticulously curated a set of high-quality biomedical primary research papers published in 2024. Our goal is to construct a dataset that accurately reflects the state-of-the-art in the field and provides a robust foundation for evaluating LLMs' capabilities for generating research ideas. Motivated by our desire to include only high-quality, peer-reviewed research, as well as those recognized by the scientific community through citations, we retrieve papers either from top venues or from other venues but are recognized by a significant number of citations. We use the Semantic Scholar API \cite{kinney2023semantic} to retrieve all biomedical papers published in top biomedical conferences according to Google Scholar venue rankings \cite{google_scholar_top_venues} in the year 2024 with at least one citation. We also retrieve papers published from other biomedical venues in the year 2024 that have at least 20 citations. Any duplicate papers are removed. We refer to these selected papers as target papers in which the ground-truth research ideas lie. 

To further enrich our dataset and provide context, we also extracted the reference papers cited by these target papers. This is done using the Semantic Scholar API as well. These reference papers contain the foundational ideas that motivated the research in the target papers, offering valuable insights into the background and rationale behind each study. By mapping each target paper to its corresponding set of reference papers, we create a comprehensive contextual framework that can aid LLMs in generating coherent and relevant research ideas. 
Also, to ensure the completeness and usability of our dataset, we disregard papers with critical missing information, such as abstracts.
This is crucial for maintaining the integrity of our evaluation, as missing information or poor contextualization could hinder LLMs in understanding the main ideas and prevent fair comparisons with generated research ideas.

\subsubsection{Relevance and Significance Based Reference Filtering} \label{sec:ref-filtering}

We believe that the reference papers provide the most significant information for generating the new research ideas in the target papers. However, not all references cited in a paper are equally relevant to its central theme. Especially when computing resources are limited, it's vital to focus on the most pertinent and significant references in the target papers. Our motivation for implementing a significance-relevancy-based filtering process is to ensure that the reference papers align closely with the target paper's primary research ideas, thus maximizing the relevance and utility of the information provided to the LLMs.
To enhance the relevance of the reference papers, we propose a filtering process that prioritizes references directly contributing to the main research idea of the target paper. This approach excludes irrelevant or overly specific references that do not align with the overarching research theme, thereby optimizing the dataset for the generation of new research ideas under constrained resources.

The filtering process is guided by three conditions:
\begin{enumerate}
    \item {\it Citation Count Threshold}. We exclude reference papers with fewer than five citations to ensure the inclusion of high-quality, widely recognized references.
    \item {\it Non-Primary Research Exclusion}. We remove non-primary research references, such as reviews, 
    editorials, letters, 
    or books, as labeled by Semantic Scholar. These sources often contain diverse ideas not directly relevant to the target paper's core research.
    \item {\it Background Section Relevance.} We also exclude reference papers that are not cited in the background section of the target paper, as they are less likely to contribute directly to the target paper's research idea.
\end{enumerate}

This filtering process ensures that the LLMs are provided with highly relevant and focused information, facilitating the generation of new and meaningful research ideas. We will use random filtering as a baseline, and the effectiveness of our filtering method will be further discussed in the ablation study section. Our approach aims to strike a balance between resource efficiency and the richness of information, thereby advancing the quality of research idea generation.

\begin{figure}[h]
\begin{tcolorbox}[colback=grey,colframe=black,title=Prompt template for generating research ideas]
You are a biomedical researcher. You are tasked with creating a hypothesis or research idea given some background knowledge. The background knowledge is provided by abstracts from other papers. \\

Here are the abstracts:\\

Abstract 1:\verb|{reference_paper_1_abstract}| \\
Abstract 2:\verb|{reference_paper_2_abstract}| \\
...... \\
Abstract n:\verb|{reference_paper_n_abstract}| \\

Using these abstracts, reason over them and come up with a novel hypothesis. Please avoid copying ideas directly, rather use the insights to inspire a novel hypothesis in the form of a brief and concise paragraph.
\end{tcolorbox}
\vskip -10pt
\caption{Prompt template used to generate research ideas.}
\label{fig:prompt_for_idea_generation}
\vskip -18pt 
\end{figure}

\subsection{Research Idea Generation}

In the process of generating a research idea, human scientists rely on relevant background information, typically reflected in the references cited in their published work. To harness the capabilities of LLMs for generating research ideas, we adopt a similar approach by grounding the LLMs in the same context considered by human researchers. Our motivation for this is to emulate human thought processes in LLMs, ensuring that the generated ideas are informed and contextually relevant. Providing LLMs with related information or context is crucial; without it, the models may struggle to meaningfully connect relevant parametric knowledge learned from their pretraining corpus.

To achieve this, 
the abstract of each target paper encapsulates the primary research idea developed by human researchers, while the abstracts of the reference papers contain the key ideas considered during the formulation of these main research ideas. For each target paper, we prompt the LLM with the abstracts of the reference papers as background information. This is accompanied by a specially designed prompt to guide the generation of new research ideas. This process is illustrated in Figure \ref{fig:prompt_for_idea_generation}, where all the \verb|{reference_paper_x_abstract}| placeholders are instantiated with the corresponding abstracts of the reference papers. 
We profile the LLMs as biomedical researchers at the beginning of the prompt. This facilitates the model's access to biomedicine-related and research-specific parametric knowledge learned from the pre-training corpus. By profiling the LLMs as biomedical researchers, we aim to maximize the utilization of the model's parametric knowledge in the biomedical domain, thereby enhancing the relevance and depth of the generated research ideas. The research ideas generated by the LLMs are then compared to the research idea presented in the target paper's abstract for evaluation.

\vspace{-0.08in}
\subsection{Evaluation of the Generated Ideas}

A straightforward approach to evaluate the quality of generated ideas is to measure the semantic similarity between the generated ideas and the idea from the target paper. However, a similarity-only metric may fail to capture the nuanced qualities of ideas generated by LLMs, such as novelty and feasibility. 
To address this, we develop a metric called the ``Insight Score", which goes beyond a similarity-only approach to assess the quality of generated ideas in a scalable and rigorous manner. The core of our metric is a personalized quality ranking which allows the users to specify any quality indicators, such as novelty, feasibility, etc. By combining personalized quality rankings with the number of generated ideas, our metric provides a nuanced measurement for various quality indicators, effectively highlighting the strengths and areas for improvement in LLMs' ability to generate research ideas. The components of our evaluation framework are detailed in the following subsections.


\subsubsection{Personalized Quality Ranking for Generated Research Ideas.}
The first step in our evaluation framework involves a personalized quality ranking. For a given target paper and reference papers pair, we first create an idea set that includes both the generated ideas and the original idea from the target paper. Details on how the original idea is extracted from the target paper are provided in the Appendix. Then we use GPT-4o to rank the quality of these ideas based on user-specified quality indicators, without revealing which idea is the original from the target paper. The motivation behind this approach is to provide a flexible and tailored assessment that aligns with the specific interests of human researchers.

The prompt template used to achieve this is shown in Figure \ref{fig:prompt_for_idea_ranking}. In the template, placeholders, denoted by curly brankets \verb|{}| allow the system to adapt to different scenarios. For instance, if a user wishes to rank research ideas based on their novelty, the system replaces \verb|{quality_indicator}| with ``novelty" in the prompt. Similarly, \verb|{target_paper_idea}| is replaced with the target paper's research idea, and \verb|{generated_idea_1}|, ... , \verb|{generated_idea_n}| are replaced with generated research ideas. The flexibility of this approach allows other quality indicators, such as feasibility, clarity, ethics, etc., to be used to rank research ideas.

Furthermore, fueled by the impressive in-context-learning ability \cite{kojima2022large} of LLMs, the system is able to accommodate a more nuanced understanding of quality indicators held in \verb|{quality_indicator}|. For example, Bob may define ``novelty" 
as ``developing new methodologies, techniques, or instruments that allow researchers to explore questions in ways that were not possible before," while Alice might consider ``novelty" as ``applying existing knowledge or technologies to address new problems or in new contexts." The system allows them to instantiate \verb|{quality_indicator}| with their respective definitions, ensuring the ranking reflects their specific interpretations. Personalized quality ranking ensures that the evaluation is aligned with the user's perspective, providing a more accurate and meaningful assessment of the generated research ideas. Additionally, by not disclosing which idea is from the target paper, the system ensures a fair and unbiased ranking of all ideas.

\begin{figure}
\begin{tcolorbox}[colback=grey,colframe=black,title=Prompt template used to rank research ideas based on user specified quality indicators]
You are a reviewer tasked with ranking the quality of a set of research ideas based on their \verb|{quality_indicator}|. The idea with the highest \verb|{quality_indicator}| should be ranked first. \\

Please rank the following hypotheses in the format:
1. Hypothesis (insert number):(insert brief rationale) \\
2. Hypothesis (insert number):(insert brief rationale) \\
3. Hypothesis (insert number):(insert brief rationale) \\
...... \\
n. Hypothesis (insert number):(insert brief rationale) \\

Please rank the following hypotheses:\\
Hypothesis 1: 
\verb|{target_paper_idea}|\\
Hypothesis 2: 
\verb|{generated_idea_1}|\\
Hypothesis 3: 
\verb|{generated_idea_2}|\\
...... \\
Hypothesis n:
\verb|{generated_idea_n}|\\

\end{tcolorbox}
\vskip -10pt
\caption{Prompt template used to rank research ideas based on user specified quality indicators.}
\label{fig:prompt_for_idea_ranking}
\vskip -14pt
\end{figure}

\subsubsection{Relative Quality Scoring for Generated Research Ideas.}

The second step in our evaluation framework is relative quality scoring, which builds upon the personalized quality ranking. The position of the target paper's idea within the ranked list of research ideas indicates the quality of the generated ideas with respect to the specified quality indicators. Intuitively, if the target paper's idea ranks higher on the list, it suggests that the generated ideas are of lower quality compared to the target paper's idea. Conversely, if the generated ideas rank higher than the target paper's idea, it indicates that the LLM is capable of producing ideas that may be of better quality than those in the target papers. To quantify different quality indicators, we introduce the following notations:

\begin{itemize}
    \item $m$: the number of target papers in our dataset.
    \item $n$: the number of research ideas an LLM generates per query.
    \item $r_{\text{target}_i}|_q$: the $i$th target paper's idea's rank within the corresponding ranked list of ideas given quality indicator $q$. When $n$ ideas are generated, $r_{\text{target}_i}|_q \in \{1, ..., n + 1\}$.
\end{itemize}

We define $I(LLM, q)$ to represent the ``Insight Score" for a given LLM based on a specific quality indicator $q$ as follows:
\vspace{-0.1in}
\begin{equation}
I(LLM, q) = \frac{1}{m} \sum_{i=1}^{m} \frac{r_{\text{target}_i}|_q - 1}{n}
\label{eq:geneius_score}
\end{equation}
Intuitively, $I(LLM, q) \in [0, 1]$. If all the target papers' ideas rank first on the list, then all the $r_{\text{target}_i}|_q = 1$, so $I(LLM, q) = 0$, indicating that the LLM is not capable of generating any research idea that surpass the quality of the target paper's idea with respect to $q$. Conversely, if all the target papers' ideas rank below all the generated ideas, that is, all the $r_{\text{target}_i}|_q = n + 1$, then $I(LLM, q) = 1$, indicating that any idea generated by the LLM is superior to the target paper's idea with respect to $q$.
Relative quality scoring provides a detailed and adaptable framework for assessing LLM performance, allowing for the consideration of user-defined quality indicators and offering insights into the model's strengths and areas for improvement.


To ensure a fair comparison across different LLMs using $I(LLM, q)$, it's important to generate the same number of research ideas $n$ for all compared LLMs. 
Our experiments show that, for a given set of target papers, the ranking of a target paper's research idea can vary depending on the number of generated ideas in the list.
This shifting of ranking positions can affect the Insight Scores of the LLMs. 
We will further discuss the effect of $n$ has on the Insight Score in the Appendix.

\section{Experiments}

\subsection{Experimental setup}
\subsubsection{Dataset.}
We curated 2,374  target papers and their corresponding 29,408 reference papers. The total number of filtered reference papers is 23,460.  
We will present the descriptive statistics of the number of references a target paper has, with and without our filtering process in the Appendix.

\begin{table*}
    \centering
    \begin{tabular}{|lc|cc|cc|cc|}
        \hline
        & \textbf{Resource}& \textbf{Semantic} & \textbf{Idea} & \textbf{Novelty} & \textbf{Feasibility} \\
        \textbf{Model} & \textbf{Scenario} & \textbf{Similarity} \textuparrow & \textbf{Overlap} \textuparrow & \textbf{Insight Score} \textuparrow & \textbf{Insight Score} \textuparrow
        \\ 

        \hline
        Llama 3.1 70B-Instruct & low  & 0.587 & 7 & 0.624 & 0.150 \\
        Llama 3.1 70B-Instruct & high  &  0.597 & \textbf{8} & 0.602 & 0.148 \\
        \hline
        Llama 3.1 405B-Instruct & low  & 0.565 & \textbf{8} & 0.647 & 0.130 \\
        Llama 3.1 405B-Instruct & high  &  0.585 & \textbf{8} & 0.677 & 0.132 \\
        \hline
        Gemini 1.5 Flash & low  & 0.585 & 7 &  0.430 & 0.242 \\
        Gemini 1.5 Flash & high  & 0.593 & \textbf{8} &  0.568 &  0.303 \\
        \hline
        Gemini 1.5 Pro & low  & 0.594 & 6 &  0.509 & \textbf{0.305} \\
        Gemini 1.5 Pro & high  & 0.604 & 7 &  0.647 &  \textbf{0.305} \\

        \hline
        GPT-3.5 Turbo & low  & 0.612 & \textbf{8} &  0.401 &  0.190 \\
        GPT-3.5 Turbo & high  & 0.619 & \textbf{8} &  0.201 &  \textbf{0.305} \\        
        \hline 
        GPT-4o Mini & low  & 0.610 & 7 &  0.446 &  0.159 \\
        GPT-4o Mini & high  & \textbf{0.620} & \textbf{8} &  0.528 & 0.207 \\
        \hline
        GPT-4o & low  &  0.599 & 7 &  0.614 &  0.143 \\
        GPT-4o & high  & 0.608 & \textbf{8} & \textbf{0.766} &  0.166 \\
        \hline

    \end{tabular}
    \caption{Main benchmark results. The table shows semantic similarity (80th percentile BERTScore (F1 score)), and the idea overlap (80th percentile LLM similarity rating) between the generated research idea and the target paper abstract, and the novelty and feasibility Insight Scores for various LLMs in high and low resource settings. Bold scores represent the highest score of a given metric.}
    \label{tab:main_results}
    \vskip -15pt
\end{table*}

\vspace{-0.05in}
\subsubsection{Models.}
To evaluate LLMs' capability of generating research ideas, we test the latest version of several most popular commercial and open-sourced LLM series with different sizes: Meta LLama Series \cite{touvron2023llama},  Google Gemini Series \cite{reid2024gemini}, and OpenAI GPT Series \cite{OpenAI_GPT4_2023}. All of these models were trained on data with cutoff dates before January 1, 2024, so the target papers published after January 1, 2024 guarantee a fair comparison by avoiding the data leakage issue.


\begin{figure}[h]
\begin{tcolorbox}[colback=grey,colframe=black,title=Prompt template used to obtain LLM similarity rating to measure the overlap of ideas between the generated research idea and the target paper.]

You are an expert in understanding and analyzing scientific content. Your task is to evaluate the degree of overlap between the ideas presented in a hypothesis and the abstract of a scientific paper. Please read both the hypothesis and the abstract carefully. Then, rate the overlap on a scale of 1 to 10, where 1 indicates minimal or no overlap, and 10 indicates a perfect or nearly perfect overlap. Provide a brief explanation for your rating. \\ 

Hypothesis: \verb|{generated_research_idea}| \\

Abstract: \verb|{target_paper_abstract}| \\

Rating:
On a scale of 1-10, rate the overlap between the ideas in the hypothesis and the abstract. \\ 

Explanation:
In one sentence, provide a brief explanation for your rating, mentioning the key points of overlap and any significant differences you observed.

\end{tcolorbox}
\vskip -10pt
\caption{Prompt template to obtain LLM similarity rating.}
\label{fig:prompt_for_llm_similarity_rating}
\vskip -18pt
\end{figure}

\vspace{-0.06in}
\subsubsection{Baseline Comparison Metrics.} 
To demonstrate the advantage of the Insight Score, we compare it with two similarity metrics: Semantic similarity and idea overlap. BERTScore (F1 score) \cite{zhang2019bertscore} is used to measure semantic similarity.
The practical upper limit of BERTScore is task dependent. To find this upper limit, we compute the BERTScore of the target papers' abstracts and their LLM-summarized research ideas and obtain an average score of 0.718. Although BERTScore ranges from 0 to 1, 0.718 is our practical upper limit. 

The LLM similarity rating, which uses GPT-4o, measures the overlap in ideas between a generated research idea and the abstract of its target paper. It outputs a rating between 0 to 10 for the overlap in ideas, along with an explanation of the rating. The prompt template for the LLM similarity rating is shown in Figure \ref{fig:prompt_for_llm_similarity_rating}.
Of the $n$ generated research ideas, the one with the highest semantic similarity is considered when measuring idea overlap.

\vspace{-0.06in}
\subsubsection{Low and High Resource Scenarios.}

To account for the high cost of inputting numerous reference paper abstracts into an LLM, we consider low and high resource scenarios to assess the capabilities of LLMs when researchers face computational constraints and when they do not.
In the low resource scenario, an LLM inputs five references filtered by our filtering method introduced earlier.
In the high resource scenario, an LLM inputs all unfiltered references, with the exception of GPT-3.5 Turbo, which truncates references that cannot fit into its context window. 
We discuss how reference filtering and the number of references affect generating research ideas in the ablation study.

\vspace{-0.06in}
\subsubsection{Different $q$  Scenarios.}

We will measure two types of quality indicators: 
feasibility and novelty. Feasibility of the ideas may be limited by the target paper because the target paper's idea has been verified by human researchers, whereas the generated ideas have not. The novelty of the generated ideas may exceed those ideas in the target papers. Given a set of reference papers, there is no guarantee that the target paper exhibits the highest level of novelty. It is possible that better 
ideas can be generated from the same set of references.

\subsection{Main Results}

We benchmark LLMs in low and high resource scenarios to assess their ability to generate research ideas. We use semantic similarity and idea overlap to measure their similarity to target papers. We also evaluate research idea generation based on two quality indicators: novelty and feasibility, using the Insight Score. In the implementation, we generate $n=3$ research ideas per query. 
The results for semantic similarity, idea overlap, and the 
novelty and feasibility Insight Scores are in Table \ref{tab:main_results}. Below we  will answer specific questions through the analysis of the results.



\subsubsection{Can LLMs generate research ideas?}
Most LLMs can generate research ideas that align well with their target papers. Table \ref{tab:main_results} shows high semantic similarity and idea overlap with target papers for most models, with GPT-4o Mini (high resource) followed by GPT-3.5 Turbo (high resource) exhibiting the highest scores. Generally, we observe that the high resource scenario generates ideas that have higher similarity scores than in the low resource scenario. These similarity scores demonstrate alignment with target paper ideas, indicating that LLMs, although they cannot see the target papers, can comprehend the background information enough to generate research ideas similar to those generated by human researchers.


\subsubsection{How well can LLMs generate novel research ideas?}
Most LLMs are capable to generate research ideas that are just as, if not, more novel than their target papers' research ideas. Any Insight Score greater than 0.5 indicates that most generated research ideas are ranked above their target papers' research ideas, concerning a quality indicator. Most of the LLMs yield novelty Insight Scores of over 0.6 with GPT-4o (high resource) having the highest score of 0.766. This means that for most LLMs, most of their generated research ideas are potentially more novel than the research idea of their target paper. 
This is significant as it demonstrates the potential of LLMs to drive scientific discovery forward with new and innovative research ideas.

\subsubsection{How well can LLMs generate feasible research ideas? }
Most LLMs generate research ideas with lower feasibility than their target papers. As shown in Table \ref{tab:main_results}, these ideas generally have low feasibility Insight Scores. GPT-3.5 Turbo (high resource) and Gemini 1.5 Pro (low and high resource) achieve the highest scores, yet all LLMs score below 0.5, indicating that most of their ideas rank lower in feasibility compared to their target papers. Although LLMs can produce novel ideas, their feasibility often remains inferior to human-generated research ideas.



\subsubsection{What is the relationship between generating novel and feasible research ideas?}
For all LLMs, there is a gap between the novelty and the feasibility of their research ideas. Table \ref{tab:main_results} shows that with the exception of GPT-3.5 Turbo (high resource), all models yield higher novelty Insight Scores than feasibility Insight Scores. The intensity of this gap varies across models. GPT-4o and the LLama 3.1 
models exhibit the largest gaps, while GPT-3.5 Turbo, GPT-4o Mini, and the Gemini series of models have smaller gaps. This indicates a general trend toward a trade-off between generating research ideas that are more novel or feasible, with the degree of the gap varying across models. This trade-off is intuitive, as research ideas that propose pursuing more novel, unexplored approaches may be less feasible to implement than ideas suggesting more incremental contributions.

\subsubsection{Can reference filtering help lower-capacity models produce more novel research ideas?} Reference filtering plays a crucial role in enabling lower-capacity models to generate more novel research ideas. As shown in Table \ref{tab:main_results}, GPT-3.5 Turbo and Llama 3.1 70B-Instruct, both smaller models in their respective families, yield higher novelty Insight Scores in the low resource scenario compared to the high resource scenario. Due to their lower capacity, these models are likely distracted by irrelevant references from target papers with fewer total references since most target papers have less than 16 references.
Thus, reference filtering becomes essential to help smaller models focus on the most relevant ideas, boosting their ability to generate more novel research ideas.

\vspace{-0.08in}
\subsection{Ablation Study}

\begin{figure}[h]
    \centering
    \includegraphics[width=1\linewidth]{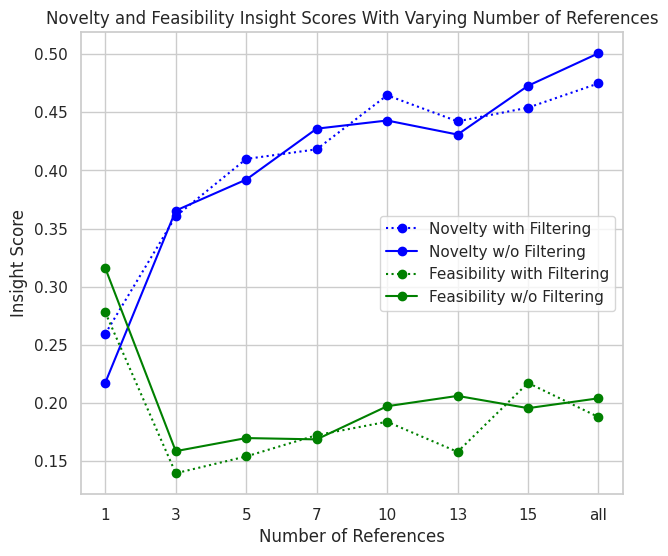}
    \vskip -10pt
    \caption{Novelty and feasibility Insight Scores as the number of filtered and unfiltered references used to generate research ideas increase.}
    \label{fig:avg_ranking-v-num_ref}
    \vskip -10pt
\end{figure}

\begin{table}[h]
\centering
\begin{tabular}{|p{0.5cm}||p{1.3cm}p{1.6cm}||p{1.4cm}p{1.6cm}||}
\hline
\textbf{\small Num Ref} & \textbf{\small Similarity (Filtered)} & \textbf{\small Similarity (Unfiltered)} & \textbf{\small Idea Overlap (Filtered)} & \textbf{\small Idea Overlap (Unfiltered)}\\
\hline
1  & \underline{0.570} & 0.567 & \underline{2.946} & 2.640 \\ \hline
3  & \underline{0.582} & 0.580 & \underline{4.636} & 4.410 \\ \hline
5  & \underline{0.586} & 0.585 & \underline{5.089} & 4.913 \\ \hline
7  & \underline{0.589} & 0.587 & \underline{5.349} & 5.304 \\ \hline
10 & 0.589 & \underline{0.590} & \underline{5.720} & 5.508 \\ \hline
13 & 0.590 & 0.590 & \underline{5.694} & 5.685 \\ \hline
15 & \underline{0.592} & 0.590 & \underline{6.002} & 5.748 \\ \hline
All & 0.592 & \underline{\textbf{0.594}} & 5.915 & \underline{\textbf{6.302}} \\ \hline
\end{tabular}
\vskip -7pt
\caption{Comparison of semantic similarity and idea overlap scores for research ideas generated by GPT-4o Mini with filtered and unfiltered references. Underlined scores are higher when compared to their filtered/unfiltered counterpart. Bold values are the highest for each measure.}
\label{table:bert_llm_comparison}
\vskip -10pt
\end{table}





For the ablation study we gathered a subset of target papers that have at least 15 references and generated research ideas using GPT-4o Mini \cite{OpenAI_GPT4_2023} with a varying number of filtered and unfiltered references. 
We evaluated the ideas generated using semantic similarity and idea overlap. 
We also calculate the Insight Scores for novelty and feasibility, which indicate the quality of the research idea with respect to novelty and feasibility. The results of these evaluations are presented in Figure \ref{fig:avg_ranking-v-num_ref} and Table \ref{table:bert_llm_comparison}. 

\subsubsection{Effectiveness of the Insight Score.}
We explore the effectiveness of our Insight Score by applying it to the quality indicators novelty and feasibility. Figure \ref{fig:avg_ranking-v-num_ref} shows how the Insight Score for novelty and feasibility evolves as we incorporate more references to generate research ideas. 
As the number of references increases, whether filtered or unfiltered, the Insight Score for novelty also increases. This shows that LLM-generated ideas tend to display more novelty when they are generated with more references. 

However, the feasibility of these LLM-generated ideas do not follow the same pattern. As shown in Figure \ref{fig:avg_ranking-v-num_ref}, increasing the number of references leads the feasibility Insight Score to plateau at a low level, regardless of whether the references are filtered. Notably, the feasibility Insight Scores remain consistently lower than the novelty Insight Scores, except in the case where the research ideas are generated with only a single reference. In this instance, the novelty Insight Score is low.

These findings demonstrate the utility of our Insight Score in capturing complex patterns that similarity metrics may overlook. 
Our results demonstrate that once a certain threshold of novelty is surpassed, the feasibility of generated ideas tends to decline and stabilize at a lower level. This observation supports the trade-off between novelty and feasibility identified in our main results, further highlighting the importance of our Insight Score in assessing the dynamics between these two important quality indicators.




\subsubsection{Effect of reference filtering on generated research ideas.} The alignment of the LLM generated research ideas to the target papers improves as the number of references increases. Specifically, filtering plays a critical role in enhancing the similarity of the generated ideas to the target paper when not all references are provided. Table \ref{table:bert_llm_comparison} shows that when all references are not available, filtered references lead to more alignment compared to unfiltered ones. Irrelevant information can cause the LLM’s output to diverge when given limited context. By filtering out less relevant references, the LLM is guided to produce ideas that are more closely aligned with the target paper.

However, when all references are available, the benefits of filtering are lost. Table \ref{table:bert_llm_comparison} shows that with all references available, unfiltered references produce the most aligned research ideas. This indicates that with sufficient references, the LLM is better equipped to ignore irrelevant information and leverage the comprehensive knowledge provided by all unfiltered references, resulting in research ideas that are most similar to the target papers.

Overall, using unfiltered references tends to produce the most aligned research ideas when all references are available. However, in scenarios with limited references, reference filtering is beneficial. This is especially relevant given the resource-intensive nature of generating ideas with LLMs as well as the input constraints of some models.

\vskip -10pt

\section{Conclusion}


In this work, we introduced \textit{IdeaBench}, a benchmark system for evaluating LLMs' ability to generate research ideas based on user-defined quality indicators. The dataset is constructed by emulating human researchers' literature review process, providing grounded contextualization for LLMs to generate research ideas. For evaluation, we proposed the ``Insight Score", a metric that surpasses similarity-based measures by capturing nuanced, user-specified quality indicators through personalized quality ranking and relative quality scoring. This work can serve as the cornerstone for academia to build up confidence in leveraging LLMs to accelerate ideation in scientific discovery.

\bibliography{aaai25}

\clearpage
\appendix

\section{Appendix}

\subsection{Code and Dataset Availability}

Due to the complexity of our dataset, we combined our dataset with all the code needed to generate our results and made them available at: \url{https://anonymous.4open.science/r/IdeaBench-2747/}

\subsection{Implementation Details}
We describe the resources used to generate and evaluate research ideas. We used various API services to generate research ideas and to evaluate them. Additionally, we employed accelerated hardware to compute semantic similarity scores between generated research ideas and their corresponding target paper abstracts.

The OpenAI API service \footnote{More details about the OpenAI API service can be found here: \url{https://platform.openai.com/docs/overview}} was employed to generate research ideas that used the OpenAI suite of models. The service was also used for extracting research ideas from target paper abstracts and evaluating research ideas with the LLM similarity rating and the Insight Score. To generate research ideas with the Gemini family of LLMs, Google AI's API service \footnote{More details about the Google AI API service can be found here: \url{https://ai.google.dev/gemini-api/docs/api-key}} was used. To generate research ideas with the Llama 3.1 family of LLMs, DeepInfra's API service \footnote{More details about the DeepInfra's API service can be found here: \url{https://deepinfra.com/}} was used. 

To evaluate the semantic similarity between research ideas and their corresponding target paper abstracts, we computed BERTScores using one NVIDIA A6000 48GB GPU. This hardware allowed for the efficient computation of BERTScores.

\subsection{Dataset Statistics}
\begin{table}[h!]
\centering
\begin{tabular}{| l | c |}
\hline
\textbf{Description} & \textbf{Count} \\
\hline
Total number of target papers & 2,374 \\
\hline
Total number of reference papers (with filtering) & 23,460 \\
\hline
Total number of reference papers (w/o filtering) & 29,408 \\
\hline
\end{tabular}
\caption{Total counts of the dataset's target papers and references. }
\label{tab:total_counts}
\end{table}
\begin{table}[h!]

\centering
\begin{tabular}{|l|c|c|}
\hline
\textbf{Statistic} & \textbf{With Filtering} & \textbf{w/o Filtering} \\
\hline
Mean & 9.882 & 12.388 \\ \hline
Standard Deviation & 6.521 & 7.946 \\ \hline
Minimum & 3 & 3 \\ \hline
25\% Percentile & 5 & 6 \\ \hline
50\% Percentile (Median) & 8 & 10 \\ \hline
75\% Percentile & 13 & 16 \\ \hline
Maximum & 51 & 62 \\ \hline
\end{tabular}
\caption{Descriptive statistics of the number of references per target paper.}
\label{tab:descriptive_stats_combined}
\vskip -10pt
\end{table}

Table \ref{tab:total_counts} shows the total number of target papers in our dataset along with the number of filtered and unfiltered references. Table \ref{tab:descriptive_stats_combined} shows the descriptive statistics of the filtered and unfiltered reference papers.

\subsection{Extracting Research Idea from a Target Paper}

\begin{figure}[h!]
\begin{tcolorbox}[colback=grey,colframe=black,title=Prompt template used to extract the research idea from a given target paper's abstract.]

Write a concise paragraph summarizing the following biomedical paper abstract as if you are proposing your own research idea or hypothesis. Focus on describing the main research idea and provide a high-level summary of the findings without detailed results or specific numerical data. Please begin the paragraph with "Hypothesis: " or "Given that ".  \\

Abstract:\\
\verb|{target_paper_abstract}|\\

Summary: 

\end{tcolorbox}
\caption{Prompt template used to extract a target paper's research idea.}
\label{fig:prompt_for_extracting_ri}
\end{figure}

To ensure a fair comparison between the research ideas generated by LLMs and those in the target paper when ranking the ideas, we extract the core research idea from the target paper’s abstract using GPT-4o with a specifically designed prompt. Abstracts often contain distracting information, such as detailed results, which may not directly reflect the central research idea and may bias the Insight Score when ranking research ideas. Therefore, we designed a prompt that focuses on summarizing the main research idea in a way that aligns with how our LLM generates ideas. Figure \ref{fig:prompt_for_extracting_ri} shows the prompt template. This process enables a fair ranking of the target paper’s idea alongside the LLM generated ideas.

\subsection{Effect of the Number of Generated Research Ideas on Insight Score}

\begin{figure} [h!] 
    \centering
    \includegraphics[width=1\linewidth]{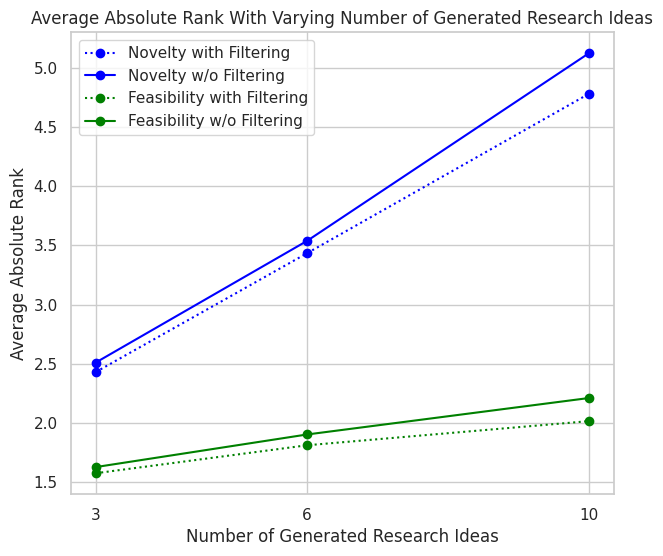}
    \caption{Effect that the number of generated research ideas has on the target paper's
    Average Absolute Rank.}
    \label{fig:avg_abs_rank_varying_n}
\end{figure}


We assess how the number of generated research ideas $n$ affects the target paper's absolute rank using GPT-4o Mini, in the same ablation study setting. Figure \ref{fig:avg_abs_rank_varying_n} shows that as $n$ increases, the target paper is ranked farther in the list for both novelty and feasibility, regardless of reference filtering. Changes in the absolute rank of the target paper will affect the Insight Score.

To illustrate the effect varying $n$ has on the Insight Score, consider we acquire the Insight Score for an LLM that generates 3 research ideas for one target paper. If the target paper ranks 3rd, then its Insight Score would be $0.667$. Now, if we have the same LLM generate 10 research ideas and the target paper ranks 5th, as suggested by Figure 
\ref{fig:avg_abs_rank_varying_n}, then its Insight Score would be $0.4$. Despite using the same LLM, the variation in $n$ causes different Insight Scores. 

Another effect that $n$ has on the Insight Score is the granularity effect, which arises from the discrete nature of the Insight Score. A larger $n$ allows for a more granular measurement of an LLM's capability in generating research ideas, meaning that a single shift in ranking position results in a smaller change in the Insight Score compared to a smaller $n$. For example, consider Case $n=6$ and Case $n=10$. If the target paper ranks around the top 50\%, that is, $r_{\text{target}_i}|_q=4$ for $n=6$ and $r_{\text{target}_i}|_q=6$ for $n=10$, both will have an Insight Score of $0.5$. However, if the target paper's ranking position drops by one, then for $n=6$, $r_{\text{target}_i}|_q=5$, the Insight Score will increase to $0.667$; whereas for $n=10$, $r_{\text{target}_i}|_q=7$, the Insight Score will be $0.6$. This difference does not necessarily indicate that the LLM with an Insight Score of $0.667$ is better than the one with $0.6$. Instead, it reflects that the first model generated fewer research ideas, resulting in a less granular Insight Score.
Thus, using the same $n$ for comparing different LLMs' Insight Scores is recommended to avoid unfair comparison.

\subsection{Case Studies}

In this section, we include a series of case studies that illustrate the capabilities of LLMs in generating research ideas, supporting our main benchmark’s findings. Each case study includes 10 examples showcasing how LLMs generate research ideas that are similar to, more novel than, and comparable in feasibility to their target papers. Additionally, two examples highlight how a smaller LLM produces incoherent and irrelevant text. Through these examples, we show the strengths of LLMs in aligning with target papers and sometimes surpassing them in novelty while maintaining comparable feasibility. We also highlight that smaller models may be incapable of generating coherent and relevant research ideas.






\subsubsection{Generated research ideas similar to their target paper.}
We find that LLMs can emulate human researchers by generating research ideas similar to those in their target papers when provided with the same background information. We present examples where LLM-generated research ideas exhibit impressive overlap with their target papers, despite these papers not being seen during the LLMs’ training. These examples are shown in Figures \ref{tab:similar_ri_1} through \ref{tab:similar_ri_10}. Each example includes the target paper’s abstract, the LLM-generated idea, and an explanation of the idea overlap rating, highlighting why the two are very similar. Key points of overlap between the target paper’s abstract and the generated research idea are spotlighted in green by human researchers.

Through these examples, we notice that LLMs are capable of identifying and leveraging the most relevant ideas from all the unfiltered references of the target papers, allowing them to generate research ideas that address issues similar to those in the target paper. Additionally, these models can produce ideas that predict findings that are related or remarkably close to those presented in the original work. This suggests that LLMs can not only identify key research questions but also anticipate the outcomes, aligning closely with the conclusions of the target papers.

\subsubsection{Novelty of the generated research ideas.}
Our main results demonstrate that LLMs can generate research ideas that are as novel, if not more so, than those in their target papers. We provide examples where LLM-generated research ideas outrank their target papers in terms of novelty in Figures \ref{tab:novel_ri_1} through \ref{tab:novel_ri_10}. These figures include the target paper’s research idea, the generated research idea, and the Insight Score’s rationale for ranking the generated idea higher in novelty. Human researchers highlight the aspects contributing to the generated idea’s novelty in green and those that illustrate why the target paper’s idea is less novel in red. These examples showcase the capability of LLMs to produce novel research ideas. 

From these examples, we observe that LLMs generate novel research ideas by creatively bridging connections between different concepts or findings. When the generated research idea is ranked higher in novelty than the target paper’s idea, it is often because the target paper’s idea incrementally builds on existing research, while the LLM-generated ideas propose new connections between different concepts or scientific findings from reference papers. This suggests that given the proper background information LLMs can generate bold and novel research ideas.

\subsubsection{Feasibility of the generated research ideas.}
Although our results show that LLMs often don't generate research ideas that are more feasible than those in their target papers, there are still some instances where LLMs’ research ideas are comparable in feasibility to their target papers' ideas. 
We provide examples where LLM-generated research ideas are comparable in feasibility to their target papers' ideas in
Figures \ref{tab:feasible_ri_1} through \ref{tab:feasible_ri_10}. These figures include the target paper’s research idea, the generated research idea, and the Insight Score’s rationale for the generated idea's feasibility. Human researchers highlight the elements that contribute to the generated idea’s feasibility in green. These examples showcase the LLMs’ ability to produce feasible research ideas. By analyzing these examples, we observe that when LLMs generate feasible research ideas, the ideas are generally straightforward and rely on established technologies and approaches.

\subsubsection{LLama 3.1 8B-Instruct results.}
In order to evaluate the effectiveness of small LLMs in generating research ideas, we tested LLama 3.1 8B-Instruct in both low-resource and high-resource conditions. The results are presented in Table \ref{tab:main_llama8b_results}. LLama 3.1 8B-Instruct reports a high idea overlap and novelty Insight Score but a low feasibility Insight Score due to generating large amounts of irrelevant or incoherent text. Our LLM-based evaluations (idea overlap and Insight Score) are not equipped to manage this because LLMs are not trained on poor quality text and our prompt templates do not consider incoherent text. As a result, the irrelevant and bad text introduces bias in our evaluations. Examples of irrelevant and incoherent text generated by LLama 3.1 8B-Instruct can be seen in Figures \ref{fig:llama_ri_1} and \ref{fig:llama_ri_2}, where sections highlighted in red by human researchers indicate problematic text produced by the LLM.


Since LLama 3.1 8B-Instruct generates a significant amount of irrelevant and nonsensical text, and our LLM evaluations are not suitable for these outputs, we opted not to include its results in the main findings. We present these outcomes to inform the community that smaller LLMs are insufficient for producing coherent research ideas.


\begin{figure*}
    \begin{minipage}{.9\textwidth}
        \centering
        \captionsetup{justification=centering}
        \begin{tcolorbox}[colback=gray!5, colframe=black, title=Similar Research Idea Example 1., fonttitle=\bfseries, width=\textwidth, center title]
            \begin{tcolorbox}[colback=white, colframe=black, title=Target Paper Abstract, fonttitle=\bfseries]
            
            Large-scale sequence modeling has sparked rapid advances that now extend into biology and genomics. However, \sethlcolor{lightgreen}\hl{modeling genomic sequences introduces challenges such as the need to model long-range token interactions, the effects of upstream and downstream regions of the genome, and the reverse complementarity (RC) of DNA.}Here, we propose an architecture motivated by these challenges that \sethlcolor{lightgreen}\hl{builds off the long-range Mamba block, and extends it to a BiMamba component that supports bi-directionality, and to a MambaDNA block that additionally supports RC equivariance.} We use MambaDNA as the basis of Caduceus, the first family of RC equivariant bi-directional long-range DNA language models, and we introduce pre-training and fine-tuning strategies that yield Caduceus DNA foundation models. \sethlcolor{lightgreen}\hl{Caduceus outperforms previous long-range models on downstream benchmarks; on a challenging long-range variant effect prediction task,} Caduceus exceeds the performance of 10x larger models that do not leverage bi-directionality or equivariance.

            \end{tcolorbox}
            \begin{tcolorbox}[colback=white, colframe=black, title=LLM Generated Research Idea, fonttitle=\bfseries]
            \textbf{LLM:} GPT-4o Mini \\
            \textbf{Resource Scenario:} High \\ 
            \textbf{Generated Research idea:} **Hypothesis:** \sethlcolor{lightgreen}\hl{We propose that leveraging the inherent reverse complement (RC) symmetry of DNA sequences can enhance the predictive accuracy of genomic models by integrating RC-equivariant architectures with state-of-the-art foundation models.} By developing a \sethlcolor{lightgreen}\hl{novel hybrid model that incorporates both the bidirectional capabilities of advanced state space models (like Mamba and S4) and the parameter-sharing techniques utilized in RC-equivariant networks}, we hypothesize that this approach will yield a more efficient and robust representation of genomic sequences. \sethlcolor{lightgreen}\hl{This model could effectively capture complex regulatory interactions and long-range dependencies in DNA}, ultimately improving the prediction of functional variants and enhancing our understanding of gene regulation mechanisms across diverse species.

            \end{tcolorbox}
            \begin{tcolorbox}[colback=white, colframe=black, title=Idea Overlap Explanation, fonttitle=\bfseries]
            
            The hypothesis and the abstract share a high degree of overlap, both focusing on leveraging the reverse complement (RC) symmetry of DNA sequences and integrating this with advanced state space models to enhance genomic predictions. Key points of overlap include the use of RC-equivariant architectures, bidirectional capabilities, and the development of a novel hybrid model to improve the prediction of functional variants. The main difference is the hypothesis's emphasis on parameter-sharing techniques, which is less explicitly mentioned in the abstract.

            \end{tcolorbox}
        \end{tcolorbox}

    \end{minipage}
    \caption{Example 1 of a generated research that is similar to its target paper. Key points of overlap between the target paper’s abstract and the generated research idea are spotlighted in green by human researchers.}
    \label{tab:similar_ri_1}
\end{figure*}

\begin{figure*}
    \begin{minipage}{.9\textwidth}
        \centering
        \captionsetup{justification=centering}
        \begin{tcolorbox}[colback=gray!5, colframe=black, title=Similar Research Idea Example 2., fonttitle=\bfseries, width=\textwidth, center title]
            \begin{tcolorbox}[colback=white, colframe=black, title=Target Paper Abstract, fonttitle=\bfseries]
            
            Traumatic brain injuries represent a leading cause of death and disability in the paediatric and adult populations. Moderate-to-severe injuries are associated with blood–brain barrier dysfunction, the development of cerebral oedema, and neuroinflammation. \sethlcolor{lightgreen}\hl{Antagonists of the tachykinin NK1 receptor have been proposed as potential agents for the post-injury treatment of TBI.} We report on the identification of EUC-001 as a potential clinical candidate for development as a novel TBI therapy. EUC-001 is a selective NK1 antagonist with a high affinity for the human NK1 receptor (Ki 5.75 × 10-10 M). It has sufficient aqueous solubility to enable intravenous administration, whilst still retaining good CNS penetration as evidenced by its ability to inhibit the gerbil foot-tapping response. \sethlcolor{lightgreen}\hl{Using an animal model of TBI, the post-injury administration of EUC-001 was shown to restore BBB function in a dose-dependent manner.EUC-001 was also able to ameliorate cerebral oedema.} These effects were associated with a significant reduction in post-TBI mortality. In addition, EUC-001 was able to significantly reduce functional deficits, both motor and cognitive, that normally follow a severe injury. EUC-001 is proposed as an ideal candidate for clinical development for TBI.

            \end{tcolorbox}
            \begin{tcolorbox}[colback=white, colframe=black, title=LLM Generated Research Idea, fonttitle=\bfseries]
            \textbf{LLM:} GPT-4o Mini \\
            \textbf{Resource Scenario:} High \\ 
            \textbf{Generated Research idea:} Based on the insights gathered from the abstracts, a compelling hypothesis emerges regarding the role of neurokinin-1 receptor (NK1R) antagonists in mitigating secondary injuries associated with traumatic brain injury (TBI). Given the established involvement of substance P (SP) in neurogenic inflammation and blood-brain barrier (BBB) disruption after CNS insults, \sethlcolor{lightgreen}\hl{it is hypothesized that early intervention with NK1R antagonists not only reduces neurogenic inflammation but also enhances BBB integrity, thereby improving outcomes in TBI patients.This dual mechanism could significantly lower the incidence of cerebral edema and excitotoxicity, leading to better functional recovery.} Future research should focus on clinical trials evaluating the timing and dosage of NK1R antagonists following TBI to establish their efficacy and safety in a therapeutic context.

            \end{tcolorbox}
            \begin{tcolorbox}[colback=white, colframe=black, title=Idea Overlap Explanation, fonttitle=\bfseries]
            
            The hypothesis and the abstract both focus on the potential benefits of NK1R antagonists in treating TBI by reducing neuroinflammation and enhancing BBB integrity. They both mention the reduction of cerebral edema and the improvement of functional outcomes. The primary difference is that the abstract specifies EUC-001 as the NK1R antagonist being studied, while the hypothesis discusses NK1R antagonists in general.

            \end{tcolorbox}
        \end{tcolorbox}

    \end{minipage}
    \caption{Example 2 of a generated research that is similar to its target paper. Key points of overlap between the target paper’s abstract and the generated research idea are spotlighted in green by human researchers.}
    \label{tab:similar_ri_2}
\end{figure*}

\begin{figure*}
    \begin{minipage}{.9\textwidth}
        \centering
        \captionsetup{justification=centering}
        \begin{tcolorbox}[colback=gray!5, colframe=black, title=Similar Research Idea Example 3., fonttitle=\bfseries, width=\textwidth, center title]
            \begin{tcolorbox}[colback=white, colframe=black, title=Target Paper Abstract, fonttitle=\bfseries]
            Patients with coronavirus disease-2019 (COVID-19) have an increased risk of thrombosis and acute respiratory distress syndrome (ARDS). Thrombosis is often attributed to increases in plasminogen activator inhibitor-1 (PAI-1) and a shut-down of fibrinolysis (blood clot dissolution). \sethlcolor{lightgreen}\hl{Decreased urokinase-type plasminogen activator (uPA), a protease necessary for cell-associated plasmin generation, and increased tissue-type plasminogen activator (tPA) and PAI-1 levels have been reported in COVID-19 patients.} Because these factors can occur in free and complexed forms with differences in their biological functions, we examined the predictive impact of uPA, tPA, and PAI-1 in their free forms and complexes as a biomarker for COVID-19 severity and the development of ARDS. In this retrospective study of 69 Japanese adults hospitalized with COVID-19 and 20 healthy donors, \sethlcolor{lightgreen}\hl{we found elevated free, non-complexed PAI-1 antigen, low circulating uPA, and uPA/PAI-1 but not tPA/PAI-1 complex levels to be associated with COVID-19 severity and ARDS development.} This biomarker profile was typical for patients in the complicated phase. Lack of PAI-1 activity in circulation despite free, non-complexed PAI-1 protein and plasmin/$\alpha$2anti-plasmin complex correlated with suPAR and sVCAM levels, markers indicating endothelial dysfunction. Furthermore, uPA/PAI-1 complex levels positively correlated with TNF$\alpha$, a cytokine reported to trigger inflammatory cell death and tissue damage. Those levels also positively correlated with lymphopenia and the pro-inflammatory factors interleukin1$\beta$ (IL1$\beta$), IL6, and C-reactive protein, markers associated with the anti-viral inflammatory response. These findings argue for using uPA and uPA/PAI-1 as novel biomarkers to detect patients at risk of developing severe COVID-19, including ARDS.

            \end{tcolorbox}
            \begin{tcolorbox}[colback=white, colframe=black, title=LLM Generated Research Idea, fonttitle=\bfseries]
            \textbf{LLM:} GPT-4o Mini \\
            \textbf{Resource Scenario:} High \\ 
            \textbf{Generated Research idea:}

            Hypothesis: \sethlcolor{lightgreen}\hl{The dysregulation of the plasminogen/plasmin system, particularly involving elevated levels of plasminogen activator inhibitor-1 (PAI-1) and urokinase plasminogen activator (uPA), plays a crucial role in the exacerbation of COVID-19-related coagulopathy and inflammation,} particularly in diverse ethnic groups. This dysregulation may lead to a hypercoagulable state that not only contributes to thrombotic events but also impairs immune responses, creating a vicious cycle of inflammation and tissue damage. \sethlcolor{lightgreen}\hl{Therefore, therapeutic strategies aimed at modulating PAI-1 and uPA levels could improve clinical outcomes in COVID-19 patients} by restoring fibrinolytic balance and enhancing the resolution of inflammation, particularly in populations at higher risk for severe disease manifestations.

            \end{tcolorbox}
            \begin{tcolorbox}[colback=white, colframe=black, title=Idea Overlap Explanation, fonttitle=\bfseries]

            Both the hypothesis and the abstract discuss the dysregulation of the plasminogen/plasmin system, particularly involving elevated PAI-1 and altered uPA levels, and their role in COVID-19-related complications such as coagulopathy, inflammation, and severe disease outcomes. The primary difference is that the hypothesis emphasizes therapeutic strategies and diverse ethnic groups, which are not addressed in the abstract.

            \end{tcolorbox}
        \end{tcolorbox}

    \end{minipage}
    \caption{Example 3 of a generated research that is similar to its target paper. Key points of overlap between the target paper’s abstract and the generated research idea are spotlighted in green by human researchers.}
    \label{tab:similar_ri_3}
    
\end{figure*}

\begin{figure*}
    \begin{minipage}{.9\textwidth}
        \centering
        \captionsetup{justification=centering}
        \begin{tcolorbox}[colback=gray!5, colframe=black, title=Similar Research Idea Example 4., fonttitle=\bfseries, width=\textwidth, center title]
            \begin{tcolorbox}[colback=white, colframe=black, title=Target Paper Abstract, fonttitle=\bfseries]
            Increased inflammation is associated with the pathogenesis of heart failure (HF). Increased circulating levels of cytokines have been previously reported and generally associated with worse clinical outcomes. In this context, the modulation of inflammation-related parameters seems to be a reasonable therapeutic option for improving the clinical course of the disease. Based on this, we aimed to compare changes in circulating cytokines when Mediterranean diet alone or in combination with hypercaloric, hyperproteic oral nutritional supplements (ONS), enriched with omega-3 (n-3) polyunsaturated fatty acids were administered to patients with HF. \sethlcolor{lightgreen}\hl{Briefly, patients were randomly assigned to receive Mediterranean Diet (control group) vs. Mediterranean Diet plus ONS (intervention group). We observed increased circulating levels of IL-6, IL-8, MCP-1 and IP-10. MCP-1 and IL-6 were associated with overweight and obesity (p = 0.01-0.01-0.04, respectively); IL-6 and IL-8 were positively correlated with fat mass and CRP serum levels (p = 0.02-0.04, respectively).} Circulating levels of IL-8 significantly decreased in all patients treated with the Mediterranean diet, while IL-6 and IP-10 only significantly decreased in patients that received plus ONS. In the univariate analysis, MCP-1 and its combination with IL-6 were associated with increased mortality (p = 0.02), while the multivariate analysis confirmed that MCP-1 was an independent factor for mortality (OR 1.01, 95\%ci 1.01-1.02). In conclusion, nutritional support using hypercaloric, hyperproteic, n-3 enriched ONS in combination with Mediterranean Diet was associated with decreased circulating levels of some cytokines and \sethlcolor{lightgreen}\hl{could represent an interesting step for improving heart functionality of patients with HF}.

            \end{tcolorbox}
            \begin{tcolorbox}[colback=white, colframe=black, title=LLM Generated Research Idea, fonttitle=\bfseries]
            \textbf{LLM:} GPT-4o Mini \\
            \textbf{Resource Scenario:} High \\ 
            \textbf{Generated Research idea:}

            Given the strong interconnections between malnutrition, sarcopenia, inflammation, and cardiovascular health highlighted across several studies, a novel hypothesis could be formulated around the \sethlcolor{lightgreen}\hl{role of targeted nutritional interventions} in modulating inflammatory responses and improving muscle mass and functionality in patients with heart failure (HF) and concurrent sarcopenia. \sethlcolor{lightgreen}\hl{Specifically, we propose that a combined supplementation of omega-3 polyunsaturated fatty acids (PUFAs) and hypercaloric, hyperproteic oral supplements enriched with leucine will significantly reduce inflammatory markers (such as IL-6 and MCP-1) while enhancing muscle mass and strength in individuals with HF and sarcopenia. This effect may subsequently improve overall cardiovascular outcomes}, including left ventricular function and quality of life, by addressing both the nutritional deficits and the inflammatory milieu that contribute to the progression of heart failure in this population. Future studies could test this hypothesis to explore the synergistic effects of these interventions on both muscle health and inflammation.

            \end{tcolorbox}
            \begin{tcolorbox}[colback=white, colframe=black, title=Idea Overlap Explanation, fonttitle=\bfseries]

            The hypothesis and abstract both focus on the role of targeted nutritional interventions, specifically hypercaloric, hyperproteic supplements enriched with omega-3 PUFAs, in reducing inflammatory markers and improving outcomes in patients with heart failure. Both documents discuss the impact on inflammatory cytokines such as IL-6 and MCP-1, and the potential improvement in heart functionality. The primary difference is that the hypothesis also emphasizes improvements in muscle mass and strength, which is not directly addressed in the abstract.

            \end{tcolorbox}
        \end{tcolorbox}

    \end{minipage}
    \caption{Example 4 of a generated research that is similar to its target paper. Key points of overlap between the target paper’s abstract and the generated research idea are spotlighted in green by human researchers.}
    \label{tab:similar_ri_4}
\end{figure*}

\begin{figure*}
    \begin{minipage}{.9\textwidth}
        \centering
        \captionsetup{justification=centering}
        \begin{tcolorbox}[colback=gray!5, colframe=black, title=Similar Research Idea Example 5., fonttitle=\bfseries, width=\textwidth, center title]
            \begin{tcolorbox}[colback=white, colframe=black, title=Target Paper Abstract, fonttitle=\bfseries]
            The Anthropocene signifies the start of a no-analogue trajectory of the Earth system that is fundamentally different from the Holocene. This new trajectory is characterized by rising risks of triggering irreversible and unmanageable shifts in Earth system functioning. \sethlcolor{lightgreen}\hl{We urgently need a new global approach to safeguard critical Earth system regulating functions more effectively and comprehensively. The global commons framework is the closest example of an existing approach with the aim of governing biophysical systems on Earth upon which the world collectively depends. Derived during stable Holocene conditions, the global commons framework must now evolve in the light of new Anthropocene dynamics}. This requires a fundamental shift from a focus only on governing shared resources beyond national jurisdiction, to one that secures critical functions of the Earth system irrespective of national boundaries. We propose a new framework—the planetary commons—which differs from the global commons framework by including not only globally shared geographic regions but also critical biophysical systems that regulate the resilience and state, and therefore livability, on Earth. \sethlcolor{lightgreen}\hl{The new planetary commons should articulate and create comprehensive stewardship obligations through Earth system governance aimed at restoring and strengthening planetary resilience and justice}.

            \end{tcolorbox}
            \begin{tcolorbox}[colback=white, colframe=black, title=LLM Generated Research Idea, fonttitle=\bfseries]
            \textbf{LLM:} GPT-3.5 Turbo \\
            \textbf{Resource Scenario:} High \\ 
            \textbf{Generated Research idea:}
             \sethlcolor{lightgreen}\hl{The increasing recognition of the Anthropocene and the urgent need for Earth system governance in the face of climate tipping points and environmental degradation suggests a shift towards a new form of international environmental law focused on active planetary stewardship}. This new paradigm of law would move beyond traditional concepts of protecting the environment to actively regulating and shaping the Earth System to ensure human and ecological well-being. \sethlcolor{lightgreen}\hl{By incorporating principles of resource equity, global justice, and planetary boundaries, this new form of international environmental law could address the challenges of the ruptured Anthropocene and promote sustainable practices on a global scale}.
            \end{tcolorbox}
            \begin{tcolorbox}[colback=white, colframe=black, title=Idea Overlap Explanation, fonttitle=\bfseries]
            Both the hypothesis and the abstract emphasize the need for a new paradigm in international environmental law or governance to address the challenges posed by the Anthropocene. They both highlight the importance of active stewardship, the incorporation of justice principles, and the necessity of managing Earth system functions to ensure planetary resilience. The primary difference is the hypothesis\'s explicit focus on "resource equity" and "global justice," which are implicit but not as explicitly stated in the abstract.

            \end{tcolorbox}
        \end{tcolorbox}

    \end{minipage}
    \caption{Example 5 of a generated research that is similar to its target paper. Key points of overlap between the target paper’s abstract and the generated research idea are spotlighted in green by human researchers.}
    \label{tab:similar_ri_5}
\end{figure*}

\begin{figure*}
    \centering
    \begin{minipage}{.9\textwidth}
        \centering
        \captionsetup{justification=centering}
        \begin{tcolorbox}[colback=gray!5, colframe=black, title=Similar Research Idea Example 6., fonttitle=\bfseries, width=\textwidth, center title]
            \begin{tcolorbox}[colback=white, colframe=black, title=Target Paper Abstract, fonttitle=\bfseries]
            The role of noncoding transcription in chromatin regulation is still controversial, extending to the \sethlcolor{lightgreen}\hl{role of transcription of antisense transcripts called COOLAIR in the Polycomb-mediated epigenetic silencing of Arabidopsis FLC (FLOWERING LOCUS C), a key step during vernalization}. Here, we show that COOLAIR transcription and PRC2 (Polycomb Repressive Complex 2) silence FLC in parallel pathways: \sethlcolor{lightgreen}\hl{an antisense-mediated transcriptional repression capable of fast response and a slow PRC2 epigenetic silencing, both of which are affected by growth dynamics and temperature fluctuations.} These features explain the varied \sethlcolor{lightgreen}\hl{importance of COOLAIR transcription} in cold-induced FLC epigenetic silencing seen in various studies using different conditions. The parallel repressive inputs and extensive feedbacks make the mechanism counterintuitive but provide great flexibility to the plant.

            \end{tcolorbox}
    
            \begin{tcolorbox}[colback=white, colframe=black, title=LLM Generated Research Idea, fonttitle=\bfseries]
            \textbf{LLM:} GPT-3.5 Turbo \\
            \textbf{Resource Scenario:} High \\ 
            \textbf{Generated Research idea:}
            Based on the insights from the abstracts provided, a novel hypothesis could be formulated to investigate the potential role of dynamic antisense transcription in regulating gene expression in response to environmental cues in plants. \sethlcolor{lightgreen}\hl{Specifically, the hypothesis could propose that the balance between sense and antisense transcription at specific gene loci, such as FLOWERING LOCUS C (FLC), is essential for coordinating the epigenetic regulation of gene expression in response to temperature fluctuations.} By manipulating the levels of \sethlcolor{lightgreen}\hl{antisense transcripts, such as COOLAIR}, and studying their impact on histone modification patterns and gene expression dynamics, this research could provide valuable insights into the molecular mechanisms underlying environmental adaptation in plants.
            \end{tcolorbox}
 
            \begin{tcolorbox}[colback=white, colframe=black, title=Idea Overlap Explanation, fonttitle=\bfseries]
            The hypothesis and the abstract exhibit a high degree of overlap. Both discuss the role of antisense transcription (specifically COOLAIR) in regulating gene expression at the FLOWERING LOCUS C (FLC) in response to environmental cues, such as temperature fluctuations. They also touch on the epigenetic regulation involving histone modifications and the potential dual pathways of transcriptional repression and epigenetic silencing. The only minor difference is that the hypothesis explicitly suggests manipulating antisense transcript levels to study their impact, which is not directly mentioned in the abstract.
            \end{tcolorbox}
            
        \end{tcolorbox}

    \end{minipage}
    \caption{Example 6 of a generated research that is similar to its target paper. Key points of overlap between the target paper’s abstract and the generated research idea are spotlighted in green by human researchers.}
    \label{tab:similar_ri_6}
\end{figure*}

\begin{figure*}
    \centering
    \begin{minipage}{.9\textwidth}
        \centering
        \captionsetup{justification=centering}
        \begin{tcolorbox}[colback=gray!5, colframe=black, title=Similar Research Idea Example 7., fonttitle=\bfseries, width=\textwidth, center title]
            \begin{tcolorbox}[colback=white, colframe=black, title=Target Paper Abstract, fonttitle=\bfseries]
            \sethlcolor{lightgreen}\hl{Salt stress profoundly affects plant growth, prompting intricate molecular responses, such as alternative splicing (AS), for environmental adaptation.} However, the response of AS events to salt stress in Opisthopappus (Opisthopappus taihangensis and Opisthopappus longilobus) remains unclear, which is a Taihang Mountain cliff-dwelling species. Using RNA-seq data, differentially expressed genes (DEGs) were identified under time and concentration gradients of salt stress. Two types of AS, skipped exon (SE) and mutually exclusive exons (MXE), were found. Differentially alternative splicing (DAS) genes in both species were significantly enriched in “protein phosphorylation”, “starch and sucrose metabolism”, and “plant hormone signal transduction” pathways. Meanwhile, distinct GO terms and KEGG pathways of DAS occurred between two species. Only a small subset of DAS genes overlapped with DEGs under salt stress. \sethlcolor{lightgreen}\hl{Although both species likely adopted protein phosphorylation to enhance salt stress tolerance, they exhibited distinct responses. The results indicated that the salt stress mechanisms of both Opisthopappus species exhibited similarities and differences in response to salt stress, which suggested that adaptive divergence might have occurred between them.} This study initially provides a comprehensive description of salt responsive AS events in Opisthopappus and conveys some insights into the molecular mechanisms behind species tolerance on the Taihang Mountains.

            \end{tcolorbox}
    
            \begin{tcolorbox}[colback=white, colframe=black, title=LLM Generated Research Idea, fonttitle=\bfseries]
            \textbf{LLM:} GPT-3.5 Turbo \\
            \textbf{Resource Scenario:} High \\ 
            \textbf{Generated Research idea:}
            Based on the information provided in the abstracts, \sethlcolor{lightgreen}\hl{a novel hypothesis could be proposed regarding the role of alternative splicing (AS) in plant response to salt stress.} Previous research has shown that AS is a widespread mechanism in plants that increases transcriptome complexity and proteome diversity, and that it can be modulated by different abiotic stresses, including heat stress and temperature fluctuations. It is known that salt stress impacts multiple aspects of plant metabolism and physiology, yet little is known about how AS may be involved in the plant\'s response to salt stress. \sethlcolor{lightgreen}\hl{Therefore, a hypothesis could be proposed that under salt stress conditions, plants undergo specific AS events that lead to the production of isoforms with altered functions, potentially playing a role in the plant\'s adaptation to saline environments.} By investigating the AS patterns in response to salt stress and identifying the genes involved in these AS events, a better understanding of the molecular mechanisms underlying salt tolerance in plants could be achieved.
            \end{tcolorbox}

            \begin{tcolorbox}[colback=white, colframe=black, title=Idea Overlap Explanation, fonttitle=\bfseries]
            The hypothesis and the abstract both focus on the role of alternative splicing (AS) in plant response to salt stress, specifically mentioning the investigation of AS patterns and identifying genes involved in these events. The key points of overlap include the emphasis on AS as a mechanism for adaptation to salt stress and the goal of understanding the molecular mechanisms underlying salt tolerance. The main difference is the specific focus on the species Opisthopappus in the abstract, which is not mentioned in the hypothesis.

            \end{tcolorbox}
            
        \end{tcolorbox}

    \end{minipage}
    \caption{Example 7 of a generated research that is similar to its target paper. Key points of overlap between the target paper’s abstract and the generated research idea are spotlighted in green by human researchers.}
    \label{tab:similar_ri_7}
\end{figure*}

\begin{figure*}
    \centering
    \begin{minipage}{.9\textwidth}
        \centering
        \captionsetup{justification=centering}
        \begin{tcolorbox}[colback=gray!5, colframe=black, title=Similar Research Idea Example 8., fonttitle=\bfseries, width=\textwidth, center title]
            \begin{tcolorbox}[colback=white, colframe=black, title=Target Paper Abstract, fonttitle=\bfseries]
            
            A role for vitamin D in immune modulation and in cancer has been suggested. In this work, we report that mice with increased availability of vitamin D display greater immune-dependent resistance to transplantable cancers and augmented responses to checkpoint blockade immunotherapies. Similarly, in humans, vitamin D–induced genes correlate with improved responses to immune checkpoint inhibitor treatment as well as with immunity to cancer and increased overall survival. \sethlcolor{lightgreen}\hl{In mice, resistance is attributable to the activity of vitamin D on intestinal epithelial cells, which alters microbiome composition in favor of Bacteroides fragilis, which positively regulates cancer immunity. Our findings indicate a previously unappreciated connection between vitamin D, microbial commensal communities, and immune responses to cancer. Collectively, they highlight vitamin D levels as a potential determinant of cancer immunity and immunotherapy success.} Editor’s summary The gut microbiome has been shown to modulate the response of cancer patients to therapy, but precisely how microbiota affect anticancer immunity is still being elucidated. Giampazolias et al. report that vitamin D bioavailability in mice influences the composition of the gut microbiome (see the Perspective by Franco and McCoy). After dietary manipulation, vitamin D levels were observed to affect gut bacteria, which in turn improved cancer immunotherapy and antitumor immunity. In humans, low vitamin D levels were correlated with tumor development, and gene signatures of vitamin D activity were associated with improved patient responses to immunotherapy. These findings highlight the connection between vitamin D and the immune system through gut bacteria and may have applications for improving cancer therapies. —Priscilla N. Kelly

            \end{tcolorbox}
    
            \begin{tcolorbox}[colback=white, colframe=black, title=LLM Generated Research Idea, fonttitle=\bfseries]
            \textbf{LLM:} GPT-3.5 Turbo \\
            \textbf{Resource Scenario:} High \\ 
            \textbf{Generated Research idea:}
            Based on the insights from the abstracts, a novel hypothesis could be that vitamin D status, particularly the levels of 25-hydroxyvitamin D, influences the composition of gut microbiota, which in turn affects the response to cancer immunotherapy. It is known that vitamin D plays a role in modulating the immune system and that gut microbiota can impact the efficacy of cancer immunotherapy. \sethlcolor{lightgreen}\hl{Therefore, it is possible that individuals with optimal vitamin D levels may have a more favorable gut microbiome composition, leading to enhanced anti-tumor immune responses and improved outcomes with cancer immunotherapy.} This hypothesis could be tested by analyzing the vitamin D levels, gut microbiota composition, and response to immunotherapy in cancer patients, to determine if there is a correlation between vitamin D status, gut microbiota diversity, and treatment outcomes.
            \end{tcolorbox}

            \begin{tcolorbox}[colback=white, colframe=black, title=Idea Overlap Explanation, fonttitle=\bfseries]
            The hypothesis and the abstract both emphasize the influence of vitamin D on gut microbiota composition and its subsequent impact on the efficacy of cancer immunotherapy. They both underscore the role of vitamin D in modulating the immune system and suggest a correlation between vitamin D levels and improved cancer treatment outcomes. The minor difference is that the abstract provides specific details about the role of Bacteroides fragilis and gene signatures of vitamin D activity, which are not mentioned in the hypothesis.

            \end{tcolorbox}
            
        \end{tcolorbox}

    \end{minipage}
    \caption{Example 8 of a generated research that is similar to its target paper. Key points of overlap between the target paper’s abstract and the generated research idea are spotlighted in green by human researchers.}
    \label{tab:similar_ri_8}
\end{figure*}

\begin{figure*}
    \centering
    \begin{minipage}{.9\textwidth}
        \centering
        \captionsetup{justification=centering}
        \begin{tcolorbox}[colback=gray!5, colframe=black, title=Similar Research Idea Example 9., fonttitle=\bfseries, width=\textwidth, center title]
            \begin{tcolorbox}[colback=white, colframe=black, title=Target Paper Abstract, fonttitle=\bfseries]
            
            Messenger RNA (mRNA) vaccines were highly effective against the ancestral SARS-CoV-2 strain, but the efficacy of bivalent mRNA boosters against XBB variants was substantially lower. \sethlcolor{lightgreen}\hl{Here, we show limited durability of neutralizing antibody (NAb) responses against XBB variants and isotype switching to immunoglobulin G4 (IgG4) responses following bivalent mRNA boosting.} Bivalent mRNA boosting elicited modest XBB.1-, XBB.1.5-, and XBB.1.16-specific NAbs that waned rapidly within 3 months. In contrast, bivalent mRNA boosting induced more robust and sustained NAbs against the ancestral WA1/2020 strain, suggesting immune imprinting. Following bivalent mRNA boosting, serum antibody responses were primarily IgG2 and IgG4 responses with poor Fc functional activity. In contrast, a third monovalent mRNA immunization boosted all isotypes including IgG1 and IgG3 with robust Fc functional activity. These data show substantial immune imprinting for the ancestral spike and isotype switching to IgG4 responses following bivalent mRNA boosting, with important implications for future booster designs and boosting strategies.

            \end{tcolorbox}
    
            \begin{tcolorbox}[colback=white, colframe=black, title=LLM Generated Research Idea, fonttitle=\bfseries]
            \textbf{LLM:} GPT-3.5 Turbo \\
            \textbf{Resource Scenario:} High \\ 
            \textbf{Generated Research idea:}
            Based on the insights from the abstracts provided, a novel hypothesis could be formulated as follows: "\sethlcolor{lightgreen}\hl{Repeated exposure to SARS-CoV-2 spike protein through mRNA vaccines leads to a shift in antibody response towards the IgG4 subclass with distinct effector functions and Fc glycosylation patterns.} This shift may impact the long-term efficacy of the vaccines by altering the balance of antibody-mediated immune responses, potentially influencing protection against infection and disease severity. Further investigation into the role of IgG4 antibodies in vaccine-induced immunity and their interplay with other IgG subclasses is warranted to optimize vaccination strategies and enhance immune protection against SARS-CoV-2 variants.
            \end{tcolorbox}

            \begin{tcolorbox}[colback=white, colframe=black, title=Idea Overlap Explanation, fonttitle=\bfseries]
            The hypothesis and the abstract have a high degree of overlap, with both discussing the shift towards IgG4 subclass responses following repeated exposure to SARS-CoV-2 spike protein via mRNA vaccines, and the implications of this shift for vaccine efficacy and immune response. Both focus on the changes in antibody subclasses, specifically IgG4, and the impact on immune protection, highlighting the need for further investigation into the role of IgG4 and other IgG subclasses in vaccine-induced immunity. The main difference is the abstract's additional focus on the comparison between bivalent and monovalent mRNA boosters, which is not explicitly addressed in the hypothesis.
            \end{tcolorbox}
            
        \end{tcolorbox}

    \end{minipage}
    \caption{Example 9 of a generated research that is similar to its target paper. Key points of overlap between the target paper’s abstract and the generated research idea are spotlighted in green by human researchers.}
    \label{tab:similar_ri_9}
\end{figure*}

\begin{figure*}
    \centering
    \begin{minipage}{.9\textwidth}
        \centering
        \captionsetup{justification=centering}
        \begin{tcolorbox}[colback=gray!5, colframe=black, title=Similar Research Idea Example 10., fonttitle=\bfseries, width=\textwidth, center title]
            \begin{tcolorbox}[colback=white, colframe=black, title=Target Paper Abstract, fonttitle=\bfseries]
            
            \sethlcolor{lightgreen}\hl{Our research has uncovered a role of cytosolic C3 in safeguarding pancreatic $\beta$-cell function during inflammation, which is crucial for diabetes. We developed an animal model with C3 selectively absent in pancreatic $\beta$-cells to demonstrate its critical importance in preserving $\beta$-cell function and protecting against inflammation-induced $\beta$-cell death.} This protective mechanism, mediated by cytosolic C3, counters the harmful effects of the proinflammatory cytokine IL-1$\beta$. These findings reveal an aspect of cytosolic C3 beyond its traditional association with the immune (complement) system, offering exciting possibilities for therapeutic interventions to preserve $\beta$--cell health and improve diabetes treatment.

            \end{tcolorbox}
    
            \begin{tcolorbox}[colback=white, colframe=black, title=LLM Generated Research Idea, fonttitle=\bfseries]
            \textbf{LLM:} GPT-3.5 Turbo \\
            \textbf{Resource Scenario:} High \\ 
            \textbf{Generated Research idea:}
            
            Based on the information provided in the abstracts, a novel hypothesis could be proposed: \sethlcolor{lightgreen}\hl{Given the evidence of the crucial role of complement C3 in promoting beta-cell survival and protecting against apoptosis in the presence of proinflammatory cytokines, it is hypothesized that manipulating the complement system, specifically targeting C3, could be a potential therapeutic strategy for enhancing beta-cell survival and function in diabetes.} By modulating the activity of C3 to enhance its prosurvival effects via AKT activation and c-Jun N-terminal kinase inhibition, it may be possible to protect beta cells from the detrimental effects of chronic inflammation and improve overall pancreatic health in diabetic individuals. Further research into the specific mechanisms by which C3 exerts its protective effects and the development of targeted interventions could lead to novel treatments for diabetes.
            
            \end{tcolorbox}
            \begin{tcolorbox}[colback=white, colframe=black, title=Idea Overlap Explanation, fonttitle=\bfseries]
            The hypothesis and the abstract exhibit a high degree of overlap, both emphasizing the critical role of complement C3 in protecting pancreatic $\beta$-cells against inflammation-induced damage and suggesting its potential as a therapeutic target for diabetes. The key points of overlap include the focus on $\beta$-cell survival, the protective role of C3 against proinflammatory cytokines, and the implications for diabetes treatment. The main difference is that the hypothesis explicitly mentions the mechanisms involving AKT activation and c-Jun N-terminal kinase inhibition, which are not detailed in the abstract.

            \end{tcolorbox} 
            
        \end{tcolorbox}

    \end{minipage}
    \caption{Example 10 of a generated research that is similar to its target paper. Key points of overlap between the target paper’s abstract and the generated research idea are spotlighted in green by human researchers.}
    \label{tab:similar_ri_10}
\end{figure*}

\begin{figure*}
    \centering
    \begin{minipage}{.9\textwidth}
        \centering
        \captionsetup{justification=centering}
        \begin{tcolorbox}[colback=gray!5, colframe=black, title=Example 1 of a Novel Generated Research Idea, fonttitle=\bfseries, width=\textwidth, center title]
            \begin{tcolorbox}[colback=white, colframe=black, title=Target Paper Research Idea, fonttitle=\bfseries]
            Given that machine learning-based scoring and classification tools like CADD are crucial for prioritizing genetic variants in various studies, \sethlcolor{lightred}\hl{the recent release of CADD v1.7 introduces advanced annotation features, including protein language model scores, regulatory variant effect predictions, and sequence conservation scores.} This latest version has been evaluated using diverse data sets and has shown enhanced performance in predicting both coding and regulatory effects. The availability of these improved scores and tools continues to support the research community in genetic variant analysis and clinical assessment. 
            \end{tcolorbox}
    
            \begin{tcolorbox}[colback=white, colframe=black, title=LLM Generated Research Idea, fonttitle=\bfseries]
            \textbf{LLM:} GPT-4o \\
            \textbf{Resource Scenario:} High \\ 
            \textbf{Generated Research idea:} Given the insights from the provided abstracts, a novel hypothesis could focus on the interplay between conserved non-coding regulatory elements and their role in human disease susceptibility, particularly within the context of evolutionary constraints and population-specific variants. **Hypothesis**: "Highly conserved non-coding cis-regulatory elements (cCREs) that exhibit evolutionary constraints across mammalian species are enriched for variants associated with complex human diseases. These variants influence disease susceptibility by modulating gene expression in a tissue-specific manner. \sethlcolor{lightgreen}\hl{Leveraging deep learning models trained on multi-species genomic alignments and functional assays of these conserved elements can improve the prediction of pathogenic non-coding variants and provide novel insights into the regulatory architecture underlying complex diseases.}" This hypothesis integrates the following insights: 1. Evolutionary conservation and constraint (Abstracts 3, 4). 2. The role of non-coding regulatory elements in gene expression and disease (Abstracts 5, 6, 26). 3. Use of deep learning and comparative genomics for predicting variant effects (Abstracts 2, 13, 29). By investigating these conserved non-coding elements and their variants, we can potentially uncover critical regulatory mechanisms contributing to disease phenotypes, thereby enhancing precision medicine approaches and therapeutic advancements.
            \end{tcolorbox}
            \begin{tcolorbox}[colback=white, colframe=black, title=Insight Score Ranking Rationale, fonttitle=\bfseries]
            1. Hypothesis B: This hypothesis \sethlcolor{lightgreen}\hl{stands out for its novel approach of integrating deep learning models with multi-species genomic alignments to predict pathogenic non-coding variants.} The focus on highly conserved non-coding cis-regulatory elements (cCREs) and their role in complex human diseases, along with leveraging evolutionary constraints, presents a unique and comprehensive method to uncover critical regulatory mechanisms ... 4. Hypothesis A: \sethlcolor{lightred}\hl{Although the advancements in CADD v1.7 are significant for genetic variant analysis, this hypothesis is more of an incremental improvement to existing tools rather than introducing a fundamentally new idea.} It focuses on the application and evaluation of an updated version of a tool, which, while useful, does not present the same level of novelty as the other hypotheses.
            \end{tcolorbox}
        \end{tcolorbox}

    \end{minipage}
    \caption{Example 1 of a novel generated research idea along with its target paper's research idea and the rationale for the ranking given by the Insight Score. Aspects contributing to the generated idea’s novelty are highlighted in green by human researchers; the reasons why the target paper’s idea is less novel are highlighted in red by human researchers.}
    \label{tab:novel_ri_1}
\end{figure*}

\begin{figure*}
    \centering
    \begin{minipage}{.9\textwidth}
        \centering
        \captionsetup{justification=centering}
        \begin{tcolorbox}[colback=gray!5, colframe=black, title=Example 2 of a Novel Generated Research Idea, fonttitle=\bfseries, width=\textwidth, center title]
            \begin{tcolorbox}[colback=white, colframe=black, title=Target Paper Research Idea, fonttitle=\bfseries]
            Given that the increasing aging population significantly strains public medical resources, exploring efficient home health care (HHC) solutions becomes critical. This research proposes addressing the service resource routing and scheduling problem by optimizing the sharing strategy among multiple HHC centers. The study aims to minimize both total operation costs and service delays. \sethlcolor{lightred}\hl{A mixed integer programming model is used to describe the problem, and a multi-objective artificial bee colony algorithm with problem-specific knowledge (MABC-PK) is developed to find solutions.}The algorithm incorporates problem-specific heuristics, a crossover operation, a self-learning neighborhood selection method, and local search techniques to refine solutions. Comparative experiments demonstrate that MABC-PK outperforms several established multi-objective optimization methods, highlighting its effectiveness for this application.
            \end{tcolorbox}
    
            \begin{tcolorbox}[colback=white, colframe=black, title=LLM Generated Research Idea, fonttitle=\bfseries]
            \textbf{LLM:} GPT-4o \\
            \textbf{Resource Scenario:} High \\ 
            \textbf{Generated Research idea:} Based on the insights gleaned from the abstracts, I propose the following novel hypothesis:"\sethlcolor{lightgreen}\hl{Integrating home healthcare logistics with smart city transportation systems using autonomous vehicles can significantly enhance the efficiency of caregiver routing and scheduling. By leveraging real-time traffic data and predictive analytics, an intelligent system can dynamically optimize routes, reduce travel times, and balance workloads among caregivers}. This approach aims to minimize operational costs while maximizing patient satisfaction and care continuity. The hypothesis can be tested by developing a hybrid optimization algorithm combining elements of vehicle routing problem (VRP), multi-objective optimization, and machine learning, and applying it to real-world home healthcare scenarios in urban environments."This hypothesis builds on the concepts of vehicle routing, scheduling optimization, autonomous vehicle coordination, and home healthcare logistics, while proposing a novel integration of these domains to address the unique challenges faced in urban home healthcare services.
            \end{tcolorbox}

            \begin{tcolorbox}[colback=white, colframe=black, title=Insight Score Ranking Rationale, fonttitle=\bfseries]
            1. Hypothesis B: \sethlcolor{lightgreen}\hl{This hypothesis stands out due to its novel integration of home healthcare logistics with smart city transportation systems using autonomous vehicles. The combination of real-time traffic data, predictive analytics, and autonomous vehicle coordination presents a unique and forward-thinking approach to optimizing caregiver routing and scheduling.} .... 4.Hypothesis A: Although this hypothesis addresses a critical problem in HHC logistics and proposes a sophisticated multi-objective artificial bee colony algorithm,\sethlcolor{lightred}\hl{it is less novel given that it essentially builds on existing optimization techniques and applies them to a specific problem without integrating new, distinct domains like the other hypotheses.}
            \end{tcolorbox}
        \end{tcolorbox}

    \end{minipage}
    \caption{Example 2 of a novel generated research idea along with its target paper's research idea and the rationale for the ranking given by the Insight Score. Aspects contributing to the generated idea’s novelty are highlighted in green by human researchers; the reasons why the target paper’s idea is less novel are highlighted in red by human researchers.}
    \label{tab:novel_ri_2}
\end{figure*}

\begin{figure*}
    \centering
    \begin{minipage}{.9\textwidth}
        \centering
        \captionsetup{justification=centering}
        \begin{tcolorbox}[colback=gray!5, colframe=black, title=Example 3 of a Novel Generated Research Idea, fonttitle=\bfseries, width=\textwidth, center title]
            \begin{tcolorbox}[colback=white, colframe=black, title=Target Paper Research Idea, fonttitle=\bfseries]
            Given that addressing sensor node and route failures in smart wireless livestock sensor networks is critical for maintaining network performance, the proposed research will investigate the time-aware UAV-based energy-efficient reconfigurable routing (TUBER) scheme. TUBER aims to mitigate the time complexity and processing cost issues noted in previous schemes by implementing a synchronized clustering-with-backup strategy, a minimum-hop neighborhood recovery mechanism, and a redundancy minimization technique. \sethlcolor{lightred}\hl{Comparative evaluations indicate that TUBER outperforms earlier schemes such as RUBER and UBER across several network performance metrics, }demonstrating significant improvements in areas including cluster survival ratio, network stability, energy dissipation, network coverage, packet delivery, fault tolerance, load balancing, routing overhead, routing delay, and failure recovery. These findings suggest that TUBER could provide a more robust and efficient solution for managing sensor node and route failures in these networks.
            \end{tcolorbox}
    
            \begin{tcolorbox}[colback=white, colframe=black, title=LLM Generated Research Idea, fonttitle=\bfseries]
            \textbf{LLM:} GPT-4o \\
            \textbf{Resource Scenario:} High \\ 
            \textbf{Generated Research idea:} Given the convergence of UAV technology, wireless sensor networks (WSNs), and advanced computational methods such as deep learning and optimization algorithms, a novel hypothesis can be formulated as follows: **Hypothesis:** Integrating UAV-based adaptive clustering with deep learning-enhanced fault-tolerant routing can significantly improve the efficiency and reliability of real-time livestock monitoring systems in large-scale, remote, and harsh environments. \sethlcolor{lightgreen}\hl{By employing UAVs equipped with high-performance computing units and multi-channel communication gateways, combined with hierarchical WSNs that utilize deep learning for predictive maintenance and fault detection, the system can dynamically optimize flight paths, enhance data collection accuracy, and extend the operational lifespan of the sensor networks.} This approach is expected to provide actionable insights into livestock health, behavior, and location, thereby improving resource allocation and management while minimizing energy consumption and system downtime. This hypothesis leverages UAV adaptability, WSN hierarchical structuring, and the robustness of deep learning models to address the challenges of extensive and unsupervised livestock farming.   
            \end{tcolorbox}

            \begin{tcolorbox}[colback=white, colframe=black, title=Insight Score Ranking Rationale, fonttitle=\bfseries]
            1. Hypothesis D: Integrating UAV-based adaptive clustering with deep learning-enhanced fault-tolerant routing is \sethlcolor{lightgreen}\hl{highly novel due to the combination of advanced computational methods like deep learning with UAV technology and WSNs for real-time livestock monitoring in harsh environments.} This approach offers a unique synergy that has the potential to significantly optimize system efficiency and reliability.... 4. Hypothesis A: While investigating the TUBER scheme for addressing sensor node and route failures in smart wireless livestock sensor networks is innovative, \sethlcolor{lightred}\hl{it primarily builds on existing schemes (RUBER and UBER)}. Although it proposes significant improvements across various network performance metrics, it is less novel compared to the integration of deep learning, edge computing, or AI-driven predictive analytics in Hypotheses B, C, and D.
            \end{tcolorbox}
        \end{tcolorbox}

    \end{minipage}
    \caption{Example 3 of a novel generated research idea along with its target paper's research idea and the rationale for the ranking given by the Insight Score. Aspects contributing to the generated idea’s novelty are highlighted in green by human researchers; the reasons why the target paper’s idea is less novel are highlighted in red by human researchers.}
    \label{tab:novel_ri_3}
\end{figure*}

\begin{figure*}
    \centering
    \begin{minipage}{.9\textwidth}
        \centering
        \captionsetup{justification=centering}
        \begin{tcolorbox}[colback=gray!5, colframe=black, title=Example 4 of a Novel Generated Research Idea, fonttitle=\bfseries, width=\textwidth, center title]
            \begin{tcolorbox}[colback=white, colframe=black, title=Target Paper Research Idea, fonttitle=\bfseries]


            Given that anesthetics impact brain regions differently depending on sex, our research aims to \sethlcolor{lightred}\hl{investigate the role of sex hormones in modulating anesthetic sensitivity}. This study reveals that females, both in mice and humans, exhibit resistance to the hypnotic effects of volatile anesthetics primarily due to the acute influence of sex hormones. Although these behavioral differences do not manifest in standard cortical electroencephalographic recordings used clinically, they are evident in subcortical activity patterns. This hidden resistance might contribute to the higher incidence of awareness during anesthesia in females, suggesting a need for sex-specific anesthetic protocols.

            \end{tcolorbox}
    
            \begin{tcolorbox}[colback=white, colframe=black, title=LLM Generated Research Idea, fonttitle=\bfseries]
            \textbf{LLM:} GPT-4o \\
            \textbf{Resource Scenario:} High \\ 
            \textbf{Generated Research idea:} 


            Based on the interplay between neural mechanisms underlying anesthesia, arousal states, and cognitive function detailed across these abstracts, I propose a novel hypothesis: "\sethlcolor{lightgreen}\hl{The differential activation and inhibition of sleep-promoting and wake-active neuronal populations within the preoptic area (POA) and their interaction with the orexinergic system are critical determinants of individual variability in cognitive recovery trajectories and the incidence of postoperative cognitive dysfunction (POCD) following general anesthesia. Specifically, dysregulation of these pathways may result in protracted neurocognitive recovery and increased susceptibility to POCD, particularly in older adults or those with preexisting neurocognitive vulnerabilities.}" This hypothesis integrates the molecular and cellular insights about POA and orexinergic neurons with the variability in anesthesia recovery patterns, aiming to elucidate the mechanisms behind cognitive resilience and dysfunction post-anesthesia.

            \end{tcolorbox}

            \begin{tcolorbox}[colback=white, colframe=black, title=Insight Score Ranking Rationale, fonttitle=\bfseries]


            1. Hypothesis B: This hypothesis stands out due to its integration of multiple novel elements: \sethlcolor{lightgreen}\hl{neural inertia, GABAergic neurons in the POA, and the neuromodulatory effects of oxytocin. The proposed interaction between these elements to explain cognitive recovery post-anesthesia is highly innovative and could lead to new therapeutic strategies.} ... 3. Hypothesis A: The hypothesis explores the role of sex hormones in modulating anesthetic sensitivity, highlighting a potential cause for higher awareness during anesthesia in females. \sethlcolor{lightred}\hl{While it is novel, the interplay between sex hormones and anesthetic effects has been previously explored to some extent}, making it slightly less novel than the first two hypotheses.

            \end{tcolorbox}
        \end{tcolorbox}

    \end{minipage}
    \caption{Example 4 of a novel generated research idea along with its target paper's research idea and the rationale for the ranking given by the Insight Score. Aspects contributing to the generated idea’s novelty are highlighted in green by human researchers; the reasons why the target paper’s idea is less novel are highlighted in red by human researchers.}
    \label{tab:novel_ri_4}
\end{figure*}

\begin{figure*}
    \centering
    \begin{minipage}{.9\textwidth}
        \centering
        \captionsetup{justification=centering}
        \begin{tcolorbox}[colback=gray!5, colframe=black, title=Example 5 of a Novel Generated Research Idea, fonttitle=\bfseries, width=\textwidth, center title]
            \begin{tcolorbox}[colback=white, colframe=black, title=Target Paper Research Idea, fonttitle=\bfseries]
            Given that detailed knowledge of synaptic circuitry in the human brain is essential for understanding brain function, this study by Shapson-Coe et al. leverages human brain tissue obtained from neurosurgical procedures to achieve a high-resolution reconstruction of a cubic millimeter of the human temporal cortex. \sethlcolor{lightred}\hl{Utilizing advanced electron microscopy and computational techniques, the study generated an extensive dataset encompassing thousands of neurons, millions of synapses, and various other brain tissue elements.} Key findings include the identification of a new class of directionally oriented neurons in deep cortical layers and the discovery of rare, powerful multisynaptic connections. The freely available dataset and analysis tools provided by the authors offer a valuable resource for future research into both normal and disordered human neural circuits.

            \end{tcolorbox}
            
            \begin{tcolorbox}[colback=white, colframe=black, title=LLM Generated Research Idea, fonttitle=\bfseries]
            \textbf{LLM:} Llama 3.1 405B-Instruct \\
            \textbf{Resource Scenario:} High \\ 
            \textbf{Generated Research idea:} The following is a novel hypothesis inspired by the abstracts provided: The human brain's ability to reorganize and adapt throughout life may be influenced by the dynamic interplay between astrocytes and neurons. Specifically, the territorial organization of astrocytes and their interactions with neurons may play a crucial role in modulating synaptic plasticity and neural circuitry. \sethlcolor{lightgreen}\hl{I propose that astrocytes act as 'synaptic editors,' selectively pruning or strengthening synapses based on neural activity patterns, thereby shaping the brain's neural networks and influencing cognitive function. Furthermore, disruptions in this astrocyte-neuron interplay may contribute to neurological disorders such as schizophrenia, where abnormalities in neural circuitry and synaptic function have been observed.} Investigating the mechanisms underlying astrocyte-mediated synaptic editing could provide new insights into brain development, plasticity, and disease." This hypothesis draws inspiration from various abstracts, including those discussing astrocyte-neuron interactions (Abstracts 8, 29), synaptic plasticity and neural circuitry (Abstracts 6, 10, 11), and neurological disorders such as schizophrenia (Abstract 33). The idea of astrocytes as 'synaptic editors' is a novel concept that integrates these different lines of research and provides a new perspective on the complex interactions between astrocytes and neurons in the brain.'
            \end{tcolorbox}

            \begin{tcolorbox}[colback=white, colframe=black, title=Insight Score Ranking Rationale, fonttitle=\bfseries]
            1. **Hypothesis C**: The idea of astrocytes acting as 'synaptic editors' is highly novel. \sethlcolor{lightgreen}\hl{This hypothesis integrates multiple lines of research to propose a new role for astrocytes in modulating synaptic plasticity and neural circuitry, as well as their potential involvement in neurological disorders. This concept of astrocytes selectively pruning or strengthening synapses based on neural activity is a fresh perspective that could lead to significant advancements in understanding brain development and function.} ... 4. **Hypothesis A**: Although this study by Shapson-Coe et al. offers a high-resolution reconstruction of the human temporal cortex and identifies new classes of neurons and synaptic connections, \sethlcolor{lightred}\hl{it is more about providing a detailed dataset and findings rather than proposing a novel hypothesis.} The discovery of new neuronal structures and connections is valuable, but it does not introduce a new conceptual framework like the other hypotheses.

            \end{tcolorbox}
        \end{tcolorbox}

    \end{minipage}
    \caption{Example 5 of a novel generated research idea along with its target paper's research idea and the rationale for the ranking given by the Insight Score. Aspects contributing to the generated idea’s novelty are highlighted in green by human researchers; the reasons why the target paper’s idea is less novel are highlighted in red by human researchers.}
    \label{tab:novel_ri_5}
\end{figure*}

\begin{figure*}
    \centering
    \begin{minipage}{.9\textwidth}
        \centering
        \captionsetup{justification=centering}
        \begin{tcolorbox}[colback=gray!5, colframe=black, title=Example 6 of a Novel Generated Research Idea, fonttitle=\bfseries, width=\textwidth, center title]
            \begin{tcolorbox}[colback=white, colframe=black, title=Target Paper Research Idea, fonttitle=\bfseries]
            
            Given that the dual public health emergencies of overdose and COVID-19 have exacerbated risks for individuals with opioid or stimulant use disorders, this study investigates the impact of Risk Mitigation Guidance (RMG) dispensations on mortality and acute care visits. Conducted as a \sethlcolor{lightred}\hl{population-based retrospective cohort study in British Columbia, Canada, the research analyzed outcomes for 5882 participants who received RMG prescriptions between March 2020 and August 2021}. The findings suggest that RMG opioid dispensations are significantly associated with reduced all-cause and overdose-related mortality, particularly with increased duration of dispensation. However, RMG stimulant dispensations did not show a significant impact on mortality, although they were linked to reduced acute care visits for any cause. This indicates that providing pharmaceutical alternatives to the illegal drug supply might be an effective intervention to lower mortality rates among individuals with opioid use disorder.

            \end{tcolorbox}
            
            \begin{tcolorbox}[colback=white, colframe=black, title=LLM Generated Research Idea, fonttitle=\bfseries]
            \textbf{LLM:} Llama 3.1 405B-Instruct \\
            \textbf{Resource Scenario:} High \\ 
            \textbf{Generated Research idea:} The abstracts discuss various aspects of substance use disorder, including treatment approaches, harm reduction strategies, and the impact of the COVID-19 pandemic on overdose rates. They also touch on the importance of addressing social determinants and using robust research methods to inform public health policy. \sethlcolor{lightgreen}\hl{A novel hypothesis that emerges from these insights is that the implementation of a community-based, peer-led harm reduction program that incorporates prescription alternatives to illicit substances, along with social support and housing assistance, may lead to a significant reduction in overdose rates and improved health outcomes among individuals with substance use disorder.} This program would prioritize the needs of marginalized populations and address the root causes of addiction, such as poverty, trauma, and lack of access to healthcare. By leveraging the expertise of people with lived experience and fostering a sense of community and connection, this approach may offer a more effective and sustainable solution to the overdose crisis than traditional treatment models.
            \end{tcolorbox}

            \begin{tcolorbox}[colback=white, colframe=black, title=Insight Score Ranking Rationale, fonttitle=\bfseries]
            1. Hypothesis C: This hypothesis proposes a novel, integrated approach to harm reduction that combines prescription alternatives to illicit substances with social support and housing assistance, led by peers with lived experience. \sethlcolor{lightgreen}\hl{It addresses multiple root causes of addiction, such as poverty and trauma, which is a novel and holistic angle not traditionally emphasized in substance use disorder treatment models.} ... 3. Hypothesis A: The investigation of Risk Mitigation Guidance (RMG) dispensations during the COVID-19 pandemic is timely and situates the study within a unique context. \sethlcolor{lightred}\hl{However, it primarily extends existing research on opioid and stimulant use disorders rather than introducing a fundamentally new concept.} ...

            \end{tcolorbox}
        \end{tcolorbox}

    \end{minipage}
    \caption{Example 6 of a novel generated research idea along with its target paper's research idea and the rationale for the ranking given by the Insight Score. Aspects contributing to the generated idea’s novelty are highlighted in green by human researchers; the reasons why the target paper’s idea is less novel are highlighted in red by human researchers.}
    \label{tab:novel_ri_6}
\end{figure*}

\begin{figure*}
    \centering
    \begin{minipage}{.9\textwidth}
        \centering
        \captionsetup{justification=centering}
        \begin{tcolorbox}[colback=gray!5, colframe=black, title=Example 7 of a Novel Generated Research Idea, fonttitle=\bfseries, width=\textwidth, center title]
            \begin{tcolorbox}[colback=white, colframe=black, title=Target Paper Research Idea, fonttitle=\bfseries]
            
            \sethlcolor{lightred}\hl{Given that nucleus pulposus cells respond differently to temporary compression (TC) and sustained compression (SC) in the context of intervertebral disc degeneration, our research aims to elucidate the mechanisms governing autophagy and apoptosis under varying compression durations.} This study shows that SC leads to severe central NP cell degeneration and heightened apoptotic activity, whereas TC results in milder degeneration with a distinct peak in autophagy markers at 6 weeks. We propose to further explore TC as a model for early autophagy-mediated degeneration and SC as a model for late-stage, apoptosis-driven degeneration. This dual approach could help identify targeted interventions for different stages of disc degeneration.

            \end{tcolorbox}
            
            \begin{tcolorbox}[colback=white, colframe=black, title=LLM Generated Research Idea, fonttitle=\bfseries]
            \textbf{LLM:} Gemini 1.5 Pro \\
            \textbf{Resource Scenario:} High \\ 
            \textbf{Generated Research idea:} Hypotheses: Given the established roles of autophagy and apoptosis in intervertebral disc degeneration (IVDD) and the emerging link between them under oxidative stress, we hypothesize that specific microRNAs (miRNAs) could regulate IVDD progression by modulating the crosstalk between these two pathways. Specifically, \sethlcolor{lightgreen}\hl{we propose that certain miRNAs, potentially upregulated in degenerative discs, could simultaneously suppress autophagy and promote apoptosis in nucleus pulposus cells, thereby exacerbating IVDD.} This hypothesis is based on the ability of miRNAs to target multiple genes within interconnected pathways and the evidence suggesting a protective role of autophagy in early IVDD stages, shifting towards a deleterious effect as apoptosis becomes dominant.

            \end{tcolorbox}

            \begin{tcolorbox}[colback=white, colframe=black, title=Insight Score Ranking Rationale, fonttitle=\bfseries]
            1. Hypothesis D: Given the established roles of autophagy and apoptosis in intervertebral disc degeneration (IVDD) and the emerging link between them under oxidative stress, we hypothesize that specific microRNAs (miRNAs) could regulate IVDD progression by modulating the crosstalk between these two pathways. \sethlcolor{lightgreen}\hl{This hypothesis is highly novel due to its focus on miRNAs as regulators of the interplay between autophagy and apoptosis, which introduces a new layer of genetic regulation and potential therapeutic targets in IVDD.} ...4. Hypothesis A: Given that nucleus pulposus cells respond differently to temporary compression (TC) and sustained compression (SC) in the context of intervertebral disc degeneration, our research aims to elucidate the mechanisms governing autophagy and apoptosis under varying compression durations. \sethlcolor{lightred}\hl{While this hypothesis is based on a dual approach to understanding compression effects, it is ranked lower in novelty because it builds on already established knowledge about the differential impacts of TC and SC, focusing on further exploration rather than introducing a fundamentally new concept}.

            \end{tcolorbox}
        \end{tcolorbox}

    \end{minipage}
    \caption{Example 7 of a novel generated research idea along with its target paper's research idea and the rationale for the ranking given by the Insight Score. Aspects contributing to the generated idea’s novelty are highlighted in green by human researchers; the reasons why the target paper’s idea is less novel are highlighted in red by human researchers.}
    \label{tab:novel_ri_7}
\end{figure*}

\begin{figure*}
    \centering
    \begin{minipage}{.9\textwidth}
        \centering
        \captionsetup{justification=centering}
        \begin{tcolorbox}[colback=gray!5, colframe=black, title=Example 8 of a Novel Generated Research Idea, fonttitle=\bfseries, width=\textwidth, center title]
            \begin{tcolorbox}[colback=white, colframe=black, title=Target Paper Research Idea, fonttitle=\bfseries]
            
            Given that insects utilize gustatory receptors (GRs) to detect various taste modalities, this study elucidates the structural mechanisms underlying sugar recognition and activation of two Drosophila sweet taste receptors, GR43a and GR64a. By using cryo-electron microscopy, the researchers revealed that both receptors form tetrameric sugar-gated cation channels, with sugar binding to ligand-binding domains inducing conformational changes that lead to channel opening. GR43a is specifically activated by the monosaccharide fructose, whereas GR64a responds to disaccharides like sucrose and maltose. \sethlcolor{lightred}\hl{These findings provide structural insights into how different sugars are recognized by insect taste receptors, offering potential targets for designing insect attractants or repellents for pest control.}

            \end{tcolorbox}
            
            \begin{tcolorbox}[colback=white, colframe=black, title=LLM Generated Research Idea, fonttitle=\bfseries]
            \textbf{LLM:} Gemini 1.5 Pro \\
            \textbf{Resource Scenario:} High \\ 
            \textbf{Generated Research idea:}
            
            Given the prevalence of CH-$\pi$ interactions in carbohydrate recognition (Abstract 3) and the importance of taste perception in mosquito behavior (Abstract 4), \sethlcolor{lightgreen}\hl{we hypothesize that disrupting CH-$\pi$ interactions between mosquito gustatory receptors and sugary food sources could lead to novel mosquito control strategies.}

            \end{tcolorbox}

            \begin{tcolorbox}[colback=white, colframe=black, title=Insight Score Ranking Rationale, fonttitle=\bfseries]
            
            1. Hypothesis B: Given the potential role of gustatory receptors like Gr43a in sensing internal nutrient levels, \sethlcolor{lightgreen}\hl{this hypothesis explores a novel and unexpected connection between taste receptors and pancreatic beta cell function, suggesting a role in type 2 diabetes. This represents a unique cross-disciplinary insight with significant implications for understanding diabetes.} ... 4. Hypothesis A: While the study of the structural mechanisms underlying sugar recognition and activation of Drosophila sweet taste receptors is valuable, \sethlcolor{lightred}\hl{it is more of an extension of existing knowledge rather than a groundbreaking new idea.} The use of cryo-electron microscopy to reveal these mechanisms is an important but expected application of this technology in structural biology.

            \end{tcolorbox}
        \end{tcolorbox}

    \end{minipage}
    \caption{Example 8 of a novel generated research idea along with its target paper's research idea and the rationale for the ranking given by the Insight Score. Aspects contributing to the generated idea’s novelty are highlighted in green by human researchers; the reasons why the target paper’s idea is less novel are highlighted in red by human researchers.}
    \label{tab:novel_ri_8}
\end{figure*}

\begin{figure*}
    \centering
    \begin{minipage}{.9\textwidth}
        \centering
        \captionsetup{justification=centering}
        \begin{tcolorbox}[colback=gray!5, colframe=black, title=Example 9 of a Novel Generated Research Idea, fonttitle=\bfseries, width=\textwidth, center title]
            \begin{tcolorbox}[colback=white, colframe=black, title=Target Paper Research Idea, fonttitle=\bfseries]
            
            Given that effectively mitigating climate change necessitates significant global behavior change, our research aims to identify the most effective strategies to motivate shifts in climate beliefs and actions. A recent study tested 11 different interventions across a large international sample and found that the interventions varied in effectiveness depending on the targeted outcome and the participants' initial climate beliefs. \sethlcolor{lightred}\hl{Specifically, interventions were moderately successful in enhancing climate beliefs and policy support among non-skeptics, but had limited impact on more effortful behaviors like tree planting.} These findings highlight the need for tailored interventions that consider the audience's predispositions and the specific behavioral goals.

            \end{tcolorbox}
            
            \begin{tcolorbox}[colback=white, colframe=black, title=LLM Generated Research Idea, fonttitle=\bfseries]
            \textbf{LLM:} Gemini 1.5 Pro \\
            \textbf{Resource Scenario:} High \\ 
            \textbf{Generated Research idea:}
            
            Hypothesis:  \sethlcolor{lightgreen}\hl{Framing climate change mitigation behaviors as collective actions that reduce psychological distance (e.g., emphasizing local community efforts with immediate and tangible benefits) will increase individual participation in pro-environmental behaviors, particularly among individuals who score high on measures of climate change helplessness.}

            \end{tcolorbox}

            \begin{tcolorbox}[colback=white, colframe=black, title=Insight Score Ranking Rationale, fonttitle=\bfseries]
            
            1. Hypothesis D: Framing climate change mitigation behaviors as collective actions that reduce psychological distance is a novel approach, particularly targeting individuals who feel helpless about climate change. \sethlcolor{lightgreen}\hl{This combines psychological concepts with community-focused interventions, offering a fresh perspective on motivating pro-environmental behaviors.} ... 4. Hypothesis A: Identifying effective strategies to motivate shifts in climate beliefs and actions based on audience predispositions and specific behavioral goals \sethlcolor{lightred}\hl{is important but more incremental. The rationale and findings suggest extensions of existing research rather than a fundamentally new approach.}

            \end{tcolorbox}
        \end{tcolorbox}

    \end{minipage}
    \caption{Example 9 of a novel generated research idea along with its target paper's research idea and the rationale for the ranking given by the Insight Score. Aspects contributing to the generated idea’s novelty are highlighted in green by human researchers; the reasons why the target paper’s idea is less novel are highlighted in red by human researchers.}
    \label{tab:novel_ri_9}
\end{figure*}

\begin{figure*}
    \centering
    \begin{minipage}{.9\textwidth}
        \centering
        \captionsetup{justification=centering}
        \begin{tcolorbox}[colback=gray!5, colframe=black, title=Example 10 of a Novel Generated Research Idea, fonttitle=\bfseries, width=\textwidth, center title]
            \begin{tcolorbox}[colback=white, colframe=black, title=Target Paper Research Idea, fonttitle=\bfseries]
            

           \sethlcolor{lightred}\hl{Given that erythroid cells rely heavily on antioxidant enzymes such as superoxide dismutase (SOD), catalase (CAT), glutathione peroxidase (GPx), and peroxiredoxin 2 (Prx2), this study explores how their mRNA levels differ in reticulocytes from healthy individuals versus those with hereditary spherocytosis (HS), sickle cell disease (SCD), and $\beta$-thalassemia ($\beta$-thal).} The investigation reveals that while reticulocyte maturity indices generally correlate with mRNA levels of these enzymes in healthy controls, this correlation is disrupted in patients with these hematologic disorders. Specifically, patients with HS, SCD, and $\beta$-thal exhibit younger reticulocytes with elevated enzyme transcript levels, albeit with different expression patterns. These findings suggest that the expression of antioxidant enzyme transcripts in reticulocytes is influenced not only by cell maturity but also by adaptive responses to abnormal erythropoiesis and altered bone marrow environments, resulting in distinct antioxidant capacities tailored to each type of anemia.

            \end{tcolorbox}
            
            \begin{tcolorbox}[colback=white, colframe=black, title=LLM Generated Research Idea, fonttitle=\bfseries]
            \textbf{LLM:} Gemini 1.5 Pro \\
            \textbf{Resource Scenario:} High \\ 
            \textbf{Generated Research idea:}
            

            Given the evidence that oxidative stress plays a significant role in sickle cell disease (SCD) pathophysiology, and that thiol modification, rather than oxidant stress alone, is crucial for phosphatidylserine (PS) exposure, we hypothesize that \sethlcolor{lightgreen}\hl{specific inhibitors of thiol oxidoreductases, enzymes responsible for thiol-disulfide exchange in proteins, could serve as a novel therapeutic target for SCD}. By selectively inhibiting these enzymes in SCD erythrocytes, we aim to reduce PS exposure, thereby mitigating hemolysis and potentially vaso-occlusive crises, without broadly affecting the cellular redox balance.

            \end{tcolorbox}

            \begin{tcolorbox}[colback=white, colframe=black, title=Insight Score Ranking Rationale, fonttitle=\bfseries]
            
            1. Hypothesis B: Given the specificity of targeting thiol oxidoreductases, this approach offers a \sethlcolor{lightgreen}\hl{novel therapeutic target focusing on thiol modification rather than broad oxidative stress, which is a new direction in SCD treatment.} ... 4. Hypothesis A: While exploring mRNA levels of antioxidant enzymes in reticulocytes from different hematologic disorders provides valuable insights, \sethlcolor{lightred}\hl{it is less novel compared to the other hypotheses, as it extends existing knowledge about erythroid cell oxidative stress response rather than introducing a new concept.}

            \end{tcolorbox}
        \end{tcolorbox}

    \end{minipage}
    \caption{Example 10 of a novel generated research idea along with its target paper's research idea and the rationale for the ranking given by the Insight Score. Aspects contributing to the generated idea’s novelty are highlighted in green by human researchers; the reasons why the target paper’s idea is less novel are highlighted in red by human researchers.}
    \label{tab:novel_ri_10}
\end{figure*}

\begin{figure*}
    \centering
    \begin{minipage}{.9\textwidth}
        \centering
        \captionsetup{justification=centering}
        \begin{tcolorbox}[colback=gray!5, colframe=black, title=Example 1 of a Feasible Generated Research Idea, fonttitle=\bfseries, width=\textwidth, center title]
            \begin{tcolorbox}[colback=white, colframe=black, title=Target Paper Research Idea, fonttitle=\bfseries]
            Given that Long Covid remains a poorly understood condition with no current diagnostic tests or therapeutic solutions, our proposed research aims to investigate the dysregulation of the complement system and its role in Long Covid. Recent multimodal proteomics analyses of blood serum from COVID-19 patients have revealed that Long Covid is characterized by persistent activation and dysregulation of the complement system, alongside elevated markers of hemolysis, tissue injury, platelet activation, and monocyte–platelet aggregates. These findings highlight the involvement of the alternative and classical complement pathways and suggest that complement-mediated thromboinflammation may contribute to the pathology of Long Covid. Our research will focus on further elucidating these pathways and exploring potential diagnostic and therapeutic interventions targeting the complement system to alleviate Long Covid symptoms.

            \end{tcolorbox}

            \begin{tcolorbox}[colback=white, colframe=black, title=LLM Generated Research Idea, fonttitle=\bfseries]
            \textbf{LLM:} Gemini 1.5 Pro \\
            \textbf{Resource Scenario:} High \\ 
            \textbf{Generated Research idea:} Given the potential link between COVID-19, complement activation, and endothelial dysfunction, we hypothesize that targeting the complement pathway, specifically the alternative pathway, \sethlcolor{lightgreen}\hl{with FDA-approved inhibitors (e.g., eculizumab)} during the acute phase of COVID-19 will reduce the incidence and severity of long COVID symptoms associated with endothelial dysfunction, such as fatigue and cognitive impairment. 
            
            \end{tcolorbox}

            \begin{tcolorbox}[colback=white, colframe=black, title=Insight Score Ranking Rationale, fonttitle=\bfseries]
            1. Hypothesis D: (Rationale: \sethlcolor{lightgreen}\hl{This hypothesis is highly feasible because it proposes using an already FDA-approved inhibitor, eculizumab, to target the complement pathway during the acute phase of COVID-19.} The specific focus on endothelial dysfunction and the use of an established drug increases its practicality and likelihood of success.) ... 3. Hypothesis A: (Rationale: This hypothesis aims to investigate the dysregulation of the complement system in Long Covid, which is a novel and complex area. While it holds significant potential for advancing understanding and treatment, the lack of current diagnostic tests or therapeutic solutions makes it less immediately feasible compared to Hypotheses C and D.)...

            \end{tcolorbox}
        \end{tcolorbox}

    \end{minipage}
    \caption{Example 1 of a feasible generated research idea along with its target paper's research idea and the rationale for the Idea Score's ranking. Elements that contribute to the generated idea’s feasibility are highlighted in green by human researchers.}
    \label{tab:feasible_ri_1}
\end{figure*}

\begin{figure*}
    \centering
    \begin{minipage}{.9\textwidth}
        \centering
        \captionsetup{justification=centering}
        \begin{tcolorbox}[colback=gray!5, colframe=black, title=Example 2 of a Feasible Generated Research Idea, fonttitle=\bfseries, width=\textwidth, center title]
            \begin{tcolorbox}[colback=white, colframe=black, title=Target Paper Research Idea, fonttitle=\bfseries]
            Given that neutrophils are integral to the tumor immune response and associated with poor clinical outcomes, our research proposes to investigate the reprogramming mechanisms that drive neutrophils into a pro-angiogenic state within the tumor microenvironment. Recent findings demonstrate that both immature and mature neutrophils infiltrating tumors undergo irreversible modifications, ultimately converging into a terminal dcTRAIL-R1+ state localized to hypoxic-glycolytic niches. These reprogrammed neutrophils enhance blood vessel formation and promote tumor growth, suggesting that targeting this pathway could improve cancer immunotherapies. Our study aims to further elucidate the ontogeny and functional trajectory of these neutrophil states across different cancer types and evaluate potential therapeutic interventions to disrupt their protumoral activities.

            \end{tcolorbox}
            
            \begin{tcolorbox}[colback=white, colframe=black, title=LLM Generated Research Idea, fonttitle=\bfseries]
            \textbf{LLM:} Gemini 1.5 Pro \\
            \textbf{Resource Scenario:} High \\ 
            \textbf{Generated Research idea:}
            Hypoxia-induced expression of CD73 on tumor-associated neutrophils promotes tumor growth and angiogenesis by increasing adenosine production and suppressing the cytotoxic activity of CD8+ T cells, suggesting that targeting the CD73-adenosine pathway in TANs may enhance the efficacy of immunotherapy. 

            \end{tcolorbox}

            \begin{tcolorbox}[colback=white, colframe=black, title=Insight Score Ranking Rationale, fonttitle=\bfseries]
            1. Hypothesis C: Hypoxia-induced expression of CD73 on tumor-associated neutrophils promotes tumor growth and angiogenesis by increasing adenosine production and suppressing the cytotoxic activity of CD8+ T cells, suggesting that targeting the CD73-adenosine pathway in TANs may enhance the efficacy of immunotherapy. (Rationale: \sethlcolor{lightgreen}\hl{This hypothesis is focused and has a clear, specific target for potential therapeutic intervention. The CD73-adenosine pathway is well-documented in literature, making this hypothesis highly feasible for investigation and clinical application.}).... 4. Hypothesis A: Given that neutrophils are integral to the tumor immune response and associated with poor clinical outcomes, our research proposes to investigate the reprogramming mechanisms that drive neutrophils into a pro-angiogenic state within the tumor microenvironment. Recent findings demonstrate that both immature and mature neutrophils infiltrating tumors undergo irreversible modifications, ultimately converging into a terminal dcTRAIL-R1+ state localized to hypoxic-glycolytic niches. These reprogrammed neutrophils enhance blood vessel formation and promote tumor growth, suggesting that targeting this pathway could improve cancer immunotherapies. Our study aims to further elucidate the ontogeny and functional trajectory of these neutrophil states across different cancer types and evaluate potential therapeutic interventions to disrupt their protumoral activities. (Rationale: This hypothesis is broad and aims to investigate multiple complex mechanisms. While scientifically interesting, its broad scope and the need for extensive preliminary work make it less feasible compared to the others.

            \end{tcolorbox}
        \end{tcolorbox}

    \end{minipage}
    \caption{Example 2 of a feasible generated research idea along with its target paper's research idea and the rationale for the Idea Score's ranking. Elements that contribute to the generated idea’s feasibility are highlighted in green by human researchers.}
    \label{tab:feasible_ri_2}
\end{figure*}

\begin{figure*}
    \centering
    \begin{minipage}{.9\textwidth}
        \centering
        \captionsetup{justification=centering}
        \begin{tcolorbox}[colback=gray!5, colframe=black, title=Example 3 of a Feasible Generated Research Idea, fonttitle=\bfseries, width=\textwidth, center title]
            \begin{tcolorbox}[colback=white, colframe=black, title=Target Paper Research Idea, fonttitle=\bfseries]
            Given that the electron transfer process (ETP) in metal single-atom catalysts (M-SACs) combined with peroxymonosulfate (PMS) systems is influenced by the electron-donating capacity of various pollutants, this research aims to explore how these characteristics affect pollutant degradation pathways. The study demonstrates that the Fenton-like activities within these systems are modulated by the electrophilic index of the pollutants, thereby altering the contributions of ETP and radical pathways. This insight could significantly enhance the understanding and optimization of SACs/PMS systems for the targeted oxidation of diverse micropollutants.

            \end{tcolorbox}
            
            \begin{tcolorbox}[colback=white, colframe=black, title=LLM Generated Research Idea, fonttitle=\bfseries]
            \textbf{LLM:} Gemini 1.5 Pro \\
            \textbf{Resource Scenario:} High \\ 
            \textbf{Generated Research idea:}
            Hypothesis: Introducing specific defects into the carbon matrix of single-atom catalysts can enhance the formation of high-valent metal-oxo species during peroxymonosulfate activation by influencing electron transfer pathways and promoting the adsorption of PMS, leading to more efficient and selective degradation of organic pollutants in water treatment. 

            \end{tcolorbox}

            \begin{tcolorbox}[colback=white, colframe=black, title=Insight Score Ranking Rationale, fonttitle=\bfseries]
            1. Hypothesis D: Introducing specific defects into the carbon matrix of single-atom catalysts can enhance the formation of high-valent metal-oxo species during peroxymonosulfate activation by influencing electron transfer pathways and promoting the adsorption of PMS, leading to more efficient and selective degradation of organic pollutants in water treatment. *Rationale:** \sethlcolor{lightgreen}\hl{This hypothesis is highly feasible as it focuses on a well-understood mechanism (defect engineering) to enhance catalytic activity. The relationship between defects and electronic structure is well-documented, and the goal of improved pollutant degradation and selectivity is well-aligned with current research trends.} ... 4. Hypothesis A: Given that the electron transfer process (ETP) in metal single-atom catalysts (M-SACs) combined with peroxymonosulfate (PMS) systems is influenced by the electron-donating capacity of various pollutants, this research aims to explore how these characteristics affect pollutant degradation pathways. The study demonstrates that the Fenton-like activities within these systems are modulated by the electrophilic index of the pollutants, thereby altering the contributions of ETP and radical pathways. **Rationale:** This hypothesis is the least feasible due to its broad scope and indirect approach. It requires a detailed understanding of the pollutants' electron-donating capacities and their electrophilic indices, which may introduce variability and complexity. The direct manipulation of the catalyst's properties (as in other hypotheses) is a more straightforward approach.

            \end{tcolorbox}
        \end{tcolorbox}

    \end{minipage}
    \caption{Example 3 of a feasible generated research idea along with its target paper's research idea and the rationale for the Idea Score's ranking. Elements that contribute to the generated idea’s feasibility are highlighted in green by human researchers.}
    \label{tab:feasible_ri_3}
\end{figure*}

\begin{figure*}
    \centering
    \begin{minipage}{.9\textwidth}
        \centering
        \captionsetup{justification=centering}
        \begin{tcolorbox}[colback=gray!5, colframe=black, title=Example 4 of a Feasible Generated Research Idea, fonttitle=\bfseries, width=\textwidth, center title]
            \begin{tcolorbox}[colback=white, colframe=black, title=Target Paper Research Idea, fonttitle=\bfseries]
            Given that the ability to control interlayer charge transport in conductive metal-organic frameworks (cMOFs) is essential for optimizing their dielectric properties, this research proposes the construction of bimetallic organic frameworks with fine-tuned interlayer spacing. By varying the ratios of Zn and Cu metal ions in the (ZnxCu3-x)(hexahydroxytriphenylene)2 (ZnCu-HHTP) system, the study achieves precise control over charge transport, band structure, and dielectric properties. The findings demonstrate that the optimized Zn3Cu1-HHTP framework exhibits superior microwave absorption, highlighting the potential of this nanotechnology-based approach for developing cMOFs with tailored interlayer characteristics for specific applications.

            \end{tcolorbox}
            
            \begin{tcolorbox}[colback=white, colframe=black, title=LLM Generated Research Idea, fonttitle=\bfseries]
            \textbf{LLM:} Gemini 1.5 Pro \\
            \textbf{Resource Scenario:} Low \\ 
            \textbf{Generated Research idea:}
            Given the insights from these abstracts, we hypothesize that incorporating conductive metal-organic frameworks (cMOFs) with tunable S···S interactions, as seen in M2(TTFTB) structures, into a flexible polymer matrix will create a material with superior electromagnetic wave absorption properties, exceeding those of cMOFs alone. This enhancement will stem from the synergistic effects of improved impedance matching due to the flexible matrix, increased conductivity from optimized S···S distances within the cMOF, and potential for "non-crosstalk" multiple perceptions due to the heterogeneous material structure. This novel composite material could pave the way for next-generation flexible electronics with integrated EMI shielding and sensing capabilities. 

            \end{tcolorbox}

            \begin{tcolorbox}[colback=white, colframe=black, title=Insight Score Ranking Rationale, fonttitle=\bfseries]
            1. Hypothesis D: \sethlcolor{lightgreen}\hl{This hypothesis is highly feasible as it combines the well-established properties of cMOFs with a flexible polymer matrix, which is a common approach in materials science. The concept of impedance matching and optimizing S···S interactions within a flexible matrix is well-supported by existing research, making this a practical and achievable goal.} ... 4. Hypothesis A: While the concept of constructing bimetallic organic frameworks with fine-tuned interlayer spacing is scientifically sound, it may be less feasible due to the challenges associated with precisely controlling the ratios of Zn and Cu metal ions and achieving the desired interlayer characteristics consistently. This approach might be more complex and less straightforward compared to the other hypotheses.")

            \end{tcolorbox}
        \end{tcolorbox}

    \end{minipage}
    \caption{Example 4 of a feasible generated research idea along with its target paper's research idea and the rationale for the Idea Score's ranking. Elements that contribute to the generated idea’s feasibility are highlighted in green by human researchers.}
    \label{tab:feasible_ri_4}
\end{figure*}

\begin{figure*}
    \centering
    \begin{minipage}{.9\textwidth}
        \centering
        \captionsetup{justification=centering}
        \begin{tcolorbox}[colback=gray!5, colframe=black, title=Example 5 of a Feasible Generated Research Idea, fonttitle=\bfseries, width=\textwidth, center title]
            \begin{tcolorbox}[colback=white, colframe=black, title=Target Paper Research Idea, fonttitle=\bfseries]
            Given that the sensitivity of liquid biopsies for detecting circulating tumor DNA (ctDNA) is limited by the scarcity of analytes in the blood, our research proposes the use of intravenous priming agents to transiently augment ctDNA levels by attenuating its clearance in vivo. This approach involves administering nanoparticles and DNA-binding antibodies 1 to 2 hours before a blood draw to inhibit cfDNA uptake by liver macrophages and protect it from nuclease degradation. Findings from preclinical cancer models indicate that these priming agents significantly increase ctDNA recovery and enhance the sensitivity of liquid biopsies, making it easier to detect small tumors and perform comprehensive tumor molecular profiling. This strategy has the potential to improve the clinical utility of liquid biopsies across various applications by boosting the detection of scarce biomarkers.

            \end{tcolorbox}
            
            \begin{tcolorbox}[colback=white, colframe=black, title=LLM Generated Research Idea, fonttitle=\bfseries]
            \textbf{LLM:} Gemini 1.5 Pro \\
            \textbf{Resource Scenario:} Low \\ 
            \textbf{Generated Research idea:} Given the limitations of current ctDNA-based cancer screening methods, particularly in sensitivity for early detection, a novel hypothesis is that combining ctDNA analysis with the detection of tumor-specific epigenetic modifications in circulating immune cells, specifically T cells, will enhance the sensitivity and specificity of early-stage cancer detection.

            \end{tcolorbox}

            \begin{tcolorbox}[colback=white, colframe=black, title=Insight Score Ranking Rationale, fonttitle=\bfseries]
            1. Hypothesis C: Combining ctDNA analysis with the detection of tumor-specific epigenetic modifications in circulating immune cells, specifically T cells, is a highly feasible approach. \sethlcolor{lightgreen}\hl{This method leverages existing ctDNA detection techniques and integrates additional biomarkers, which can be validated through established laboratory methods. The use of T cells for epigenetic analysis is also supported by advancements in immunology and can be achieved with current technologies}. ... 4. Hypothesis A: Using intravenous priming agents to augment ctDNA levels by attenuating its clearance is the least feasible. This strategy involves novel and complex interventions, including the development of nanoparticles and DNA-binding antibodies, and requires careful tuning of timing and dosage. The potential for adverse effects and the need for rigorous validation in preclinical and clinical settings make this approach the most challenging to implement.

            \end{tcolorbox}
        \end{tcolorbox}

    \end{minipage}
    \caption{Example 5 of a feasible generated research idea along with its target paper's research idea and the rationale for the Idea Score's ranking. Elements that contribute to the generated idea’s feasibility are highlighted in green by human researchers.}
    \label{tab:feasible_ri_5}
\end{figure*}

\begin{figure*}
    \centering
    \begin{minipage}{.9\textwidth}
        \centering
        \captionsetup{justification=centering}
        \begin{tcolorbox}[colback=gray!5, colframe=black, title=Example 6 of a Feasible Generated Research Idea, fonttitle=\bfseries, width=\textwidth, center title]
            \begin{tcolorbox}[colback=white, colframe=black, title=Target Paper Research Idea, fonttitle=\bfseries]
            Given that optimizing frictional behavior in devices like touchscreens and robotic hands is crucial yet challenging due to the lack of systematic methods to design dry contact interfaces with specific friction laws, this study proposes a novel surface design strategy. The proposed method simplifies the multiscale problem of tribology by treating surface topographies as assemblies of spherical asperities, allowing precise control over the frictional properties by adjusting the asperities' heights. By demonstrating various elastomer-glass metainterfaces with both linear and nonlinear friction laws, this approach offers a scalable, material-independent, and chemical-free solution for creating energy-efficient and adaptable smart interfaces. This design strategy holds promise for broad applicability across different material combinations and scales, enabling the pre-determined customization of frictional relationships.

            \end{tcolorbox}
            
            \begin{tcolorbox}[colback=white, colframe=black, title=LLM Generated Research Idea, fonttitle=\bfseries]
            \textbf{LLM:} Gemini 1.5 Pro \\
            \textbf{Resource Scenario:} Low \\ 
            \textbf{Generated Research idea:} Building on the findings that both microscopic junction behavior and macroscopic shoe properties influence friction dynamics, we hypothesize that strategically engineering the surface texture of athletic shoe outsoles, inspired by the variability of microscopic junction behavior, can allow for targeted manipulation of friction coefficients, potentially optimizing grip during specific athletic maneuvers like pivoting or accelerating.

            \end{tcolorbox}

            \begin{tcolorbox}[colback=white, colframe=black, title=Insight Score Ranking Rationale, fonttitle=\bfseries]

            1. Hypothesis B: (This hypothesis leverages well-understood principles of friction dynamics and applies them to the practical and well-defined problem of optimizing athletic shoe outsoles. \sethlcolor{lightgreen}\hl{The approach is relatively straightforward, focusing on the surface texture of outsoles, which is a feasible engineering challenge.)} ... 3. Hypothesis A: (This hypothesis presents a novel and systematic method for designing dry contact interfaces with specific friction laws by adjusting asperities' heights. While innovative and promising, it involves a more complex multiscale problem and requires precise control over surface topographies, which may present more significant feasibility challenges.)

            \end{tcolorbox}
        \end{tcolorbox}

    \end{minipage}
    \caption{Example 6 of a feasible generated research idea along with its target paper's research idea and the rationale for the Idea Score's ranking. Elements that contribute to the generated idea’s feasibility are highlighted in green by human researchers.}
    \label{tab:feasible_ri_6}
\end{figure*}

\begin{figure*}
    \centering
    \begin{minipage}{.9\textwidth}
        \centering
        \captionsetup{justification=centering}
        \begin{tcolorbox}[colback=gray!5, colframe=black, title=Example 7 of a Feasible Generated Research Idea, fonttitle=\bfseries, width=\textwidth, center title]
            \begin{tcolorbox}[colback=white, colframe=black, title=Target Paper Research Idea, fonttitle=\bfseries]
            Given that the current ecological theory, particularly models based on the generalized Lotka-Volterra (GLV) framework, predicts that increased species diversity leads to instability, there is a compelling need to reconcile this with observations of stable, biodiverse ecosystems such as tropical rainforests and coral reefs. This research proposes that modeling population growth as a sublinear power law, rather than logistic growth, resolves the diversity-stability debate by demonstrating that diversity actually promotes stability. The sublinear growth model suggests that competitive interactions among populations do not lead to exclusion but instead enhance stability as diversity increases. This theory aligns with empirical macroecological patterns and predicts that biodiversity loss could accelerate ecosystem destabilization. Therefore, adopting a sublinear growth model could provide a more accurate theoretical framework for understanding and preserving ecosystem stability in the face of biodiversity loss.

            \end{tcolorbox}
            
            \begin{tcolorbox}[colback=white, colframe=black, title=LLM Generated Research Idea, fonttitle=\bfseries]
            \textbf{LLM:} Gemini 1.5 Pro \\
            \textbf{Resource Scenario:} Low \\ 
            \textbf{Generated Research idea:} Given the insights from these abstracts, a novel hypothesis emerges: **The stability and diversity of ecological communities are significantly influenced by the degree of metabolic similarity between species, with highly similar metabolic networks leading to increased competition for limiting resources and decreased stability, particularly under resource scarcity.** This hypothesis integrates the concepts of resource competition, metabolic constraints (like Liebig's Law), and the influence of resource availability on community dynamics. It suggests that species with highly overlapping metabolic needs will experience intensified competition for the same limiting resource(s), potentially leading to exclusion or unstable coexistence. Conversely, communities with greater metabolic diversity may exhibit greater stability due to niche partitioning and reduced competition for limiting resources. This hypothesis can be tested by constructing experimental communities with varying degrees of metabolic overlap and monitoring their stability and dynamics under different resource conditions.

            \end{tcolorbox}

            \begin{tcolorbox}[colback=white, colframe=black, title=Insight Score Ranking Rationale, fonttitle=\bfseries]
            1. Hypothesis B: The hypothesis that "the stability and diversity of ecological communities are significantly influenced by the degree of metabolic similarity between species" is highly feasible. \sethlcolor{lightgreen}\hl{This idea can be empirically tested by constructing experimental communities and measuring stability and competitive dynamics, which is a relatively straightforward and controlled approach to validate the hypothesis.} ... 4. Hypothesis A: While proposing a sublinear growth model to reconcile the diversity-stability debate is theoretically interesting and aligns with macroecological patterns, it may be less immediately feasible. This approach involves complex mathematical modeling and may require extensive empirical data across various ecosystems to substantiate the model, making it more challenging to test and validate in the short term.')

            \end{tcolorbox}
        \end{tcolorbox}

    \end{minipage}
    \caption{Example 7 of a feasible generated research idea along with its target paper's research idea and the rationale for the Idea Score's ranking. Elements that contribute to the generated idea’s feasibility are highlighted in green by human researchers.}
    \label{tab:feasible_ri_7}
\end{figure*}

\begin{figure*}
    \centering
    \begin{minipage}{.9\textwidth}
        \centering
        \captionsetup{justification=centering}
        \begin{tcolorbox}[colback=gray!5, colframe=black, title=Example 8 of a Feasible Generated Research Idea, fonttitle=\bfseries, width=\textwidth, center title]
            \begin{tcolorbox}[colback=white, colframe=black, title=Target Paper Research Idea, fonttitle=\bfseries]
            Given that the form and function of biomolecular condensates are intricately linked to their material properties, we propose to investigate the physical determinants of condensate fluid phase dynamics by integrating microrheology with molecular simulations. Our preliminary findings suggest that the dynamical properties of condensates, such as mechanical relaxation times and energy barriers for network reconfiguration, are influenced by factors like intermolecular interactions, binding energy, and chain length. Specifically, while mechanical relaxation times are dependent on both intermolecular interactions and chain length, the flow activation energy is solely governed by the strengths of intermolecular interactions. This indicates a complex interplay between viscoelasticity and flow activation energy in biomolecular diffusion within condensates. Understanding these distinctive roles could provide deeper insights into the material and transport properties of biomolecular condensates, potentially leading to novel biomedical applications.

            \end{tcolorbox}
            
            \begin{tcolorbox}[colback=white, colframe=black, title=LLM Generated Research Idea, fonttitle=\bfseries]
            \textbf{LLM:} Gemini 1.5 Pro \\
            \textbf{Resource Scenario:} Low \\ 
            \textbf{Generated Research idea:} Given the emerging understanding of how short-range interactions (like cation-$\pi$) and long-range interactions (like electrostatic forces) interplay in biomolecular condensate formation, we hypothesize that modulating the sequence-specific distribution of charged and aromatic residues within intrinsically disordered regions of condensate-forming proteins can be used to fine-tune the viscosity and transport properties of these condensates. Specifically, we propose that increasing the local clustering of cationic and aromatic residues will enhance cation-$\pi$ interactions, leading to decreased diffusion rates and increased viscosity within the condensate, without significantly altering the overall phase separation propensity. This hypothesis could be tested by systematically engineering synthetic proteins with varying arrangements of charged and aromatic residues and characterizing their phase behavior, condensate viscosity, and molecular diffusion rates using techniques like fluorescence correlation spectroscopy and microrheology.

            \end{tcolorbox}

            \begin{tcolorbox}[colback=white, colframe=black, title=Insight Score Ranking Rationale, fonttitle=\bfseries]
            
            1. Hypothesis D: Given the \sethlcolor{lightgreen}\hl{clear and direct approach of modulating specific protein sequences to alter condensate properties, this hypothesis appears highly feasible. The use of synthetic proteins and well-established techniques like fluorescence correlation spectroscopy and microrheology provides a straightforward and practical experimental pathway.} ... 3. Hypothesis A: Investigating the physical determinants of condensate fluid phase dynamics through microrheology and molecular simulations is feasible, but it involves significant complexity in integrating and interpreting data from multiple advanced techniques. While promising, it may require substantial preliminary work to establish robust models and experimental protocols. ...

            \end{tcolorbox}
        \end{tcolorbox}

    \end{minipage}
    \caption{Example 8 of a feasible generated research idea along with its target paper's research idea and the rationale for the Idea Score's ranking. Elements that contribute to the generated idea’s feasibility are highlighted in green by human researchers.}
    \label{tab:feasible_ri_8}
\end{figure*}

\begin{figure*}
    \centering
    \begin{minipage}{.9\textwidth}
        \centering
        \captionsetup{justification=centering}
        \begin{tcolorbox}[colback=gray!5, colframe=black, title=Example 9 of a Feasible Generated Research Idea, fonttitle=\bfseries, width=\textwidth, center title]
            \begin{tcolorbox}[colback=white, colframe=black, title=Target Paper Research Idea, fonttitle=\bfseries]
            
            Given that urban vegetation has been traditionally promoted for its potential to improve air quality, our research aims to explore the nuanced interactions between urban greening and air pollution at different scales. This study reveals that while urban vegetation offers various health benefits, its effectiveness in mitigating air pollution is limited and may even worsen conditions at the street level. Consequently, our research will focus on developing strategies to reduce anthropogenic emissions as a more effective approach to enhancing urban air quality.

            \end{tcolorbox}
            
            \begin{tcolorbox}[colback=white, colframe=black, title=LLM Generated Research Idea, fonttitle=\bfseries]
            \textbf{LLM:} Gemini 1.5 Flash \\
            \textbf{Resource Scenario:} High \\ 
            \textbf{Generated Research idea:} 

            Hypothesis: 
            Urban green spaces, specifically those dominated by native tree species, may mitigate the adverse health effects of air pollution by acting as a natural filter for particulate matter (PM2.5) and nitrogen dioxide (NO2) while simultaneously reducing the allergenic potential of pollen. This effect is particularly important in urban areas with high traffic density and existing vegetation that contributes to aeroallergen production. This hypothesis posits that carefully planned urban tree planting initiatives, considering both air quality and allergenic properties of tree species, could lead to a significant reduction in asthma and allergy-related health issues in urban populations.

            \end{tcolorbox}

            \begin{tcolorbox}[colback=white, colframe=black, title=Insight Score Ranking Rationale, fonttitle=\bfseries]
            
            1. Hypothesis B: The hypothesis focuses on the specific tree species composition and local meteorological conditions, making it a highly feasible and targeted approach. \sethlcolor{lightgreen}\hl{Investigating the impact of different tree species on local microclimates and air quality under varying conditions is a practical and measurable research endeavor that can provide actionable insights for urban planning.} ... 4. Hypothesis A: While this hypothesis is grounded in the reality that urban vegetation has limited effectiveness in certain contexts, its feasibility is lower because it involves the broad goal of developing new strategies to reduce anthropogenic emissions. This is a more complex and long-term endeavor compared to the more focused and specific studies proposed in the other hypotheses.

            \end{tcolorbox}
        \end{tcolorbox}

    \end{minipage}
    \caption{Example 9 of a feasible generated research idea along with its target paper's research idea and the rationale for the Idea Score's ranking. Elements that contribute to the generated idea’s feasibility are highlighted in green by human researchers.}
    \label{tab:feasible_ri_9}
\end{figure*}

\begin{figure*}
    \centering
    \begin{minipage}{.9\textwidth}
        \centering
        \captionsetup{justification=centering}
        \begin{tcolorbox}[colback=gray!5, colframe=black, title=Example 10 of a Feasible Generated Research Idea, fonttitle=\bfseries, width=\textwidth, center title]
            \begin{tcolorbox}[colback=white, colframe=black, title=Target Paper Research Idea, fonttitle=\bfseries]
            
            Given that the phase III CheckMate 722 trial evaluated the efficacy of nivolumab combined with chemotherapy compared to chemotherapy alone in patients with EGFR-mutated metastatic non–small-cell lung cancer (NSCLC) who had progressed after EGFR TKI therapy, the study found that progression-free survival (PFS) was not significantly improved with the addition of nivolumab. However, subgroup analyses indicated a trend favoring the combination therapy in patients with sensitizing EGFR mutations or those who had received only one line of prior EGFR TKI treatment. Overall survival (OS) and objective response rates (ORR) showed modest improvements, although the combination therapy was associated with higher rates of grade 3/4 treatment-related adverse events. This suggests that while nivolumab plus chemotherapy may offer some benefits in specific patient subgroups, the overall efficacy and safety profile warrant further investigation.

            \end{tcolorbox}
            
            \begin{tcolorbox}[colback=white, colframe=black, title=LLM Generated Research Idea, fonttitle=\bfseries]
            \textbf{LLM:} Gemini 1.5 Flash \\
            \textbf{Resource Scenario:} High \\ 
            \textbf{Generated Research idea:} 

            Hypothesis: 
            Given the observed increase in PD-L1 expression following EGFR-TKI treatment in EGFR-mutant NSCLC (Abstract 8), and the demonstrated efficacy of immunotherapy in this setting (Abstract 4), a hypothesis emerges: **Combining immunotherapy with EGFR-TKIs in patients with EGFR-mutant NSCLC who have progressed on EGFR-TKI monotherapy could improve treatment outcomes.** This approach could potentially overcome the acquired resistance associated with EGFR-TKI treatment by leveraging the immune system to target the tumor cells. This hypothesis is particularly relevant considering the growing evidence for TMB as a biomarker for immunotherapy response in NSCLC (Abstract 11), and the potential for EGFR-TKIs to alter the tumor microenvironment in a way that favors immunotherapy (Abstract 8). 

            \end{tcolorbox}

            \begin{tcolorbox}[colback=white, colframe=black, title=Insight Score Ranking Rationale, fonttitle=\bfseries]
            
            1. Hypothesis B: Combining immunotherapy with EGFR-TKIs in patients with EGFR-mutant NSCLC who have developed resistance to EGFR-TKIs alone could improve clinical outcomes compared to EGFR-TKI monotherapy. \sethlcolor{lightgreen}\hl{(This hypothesis is based on leveraging increased PD-L1 expression after EGFR-TKI treatment, which is a well-documented phenomenon and has tangible mechanistic rationale and existing preliminary evidence supporting it.)} ... 4. Hypothesis A: Nivolumab combined with chemotherapy may offer some benefits in specific patient subgroups with EGFR-mutant NSCLC, but overall efficacy and safety profile warrant further investigation. (This hypothesis is based on existing clinical trial data but has shown only modest improvements with higher adverse events, making it less feasible compared to the other hypotheses which propose novel and potentially more effective combinations.

            \end{tcolorbox}
        \end{tcolorbox}

    \end{minipage}
    \caption{Example 10 of a feasible generated research idea along with its target paper's research idea and the rationale for the Idea Score's ranking. Elements that contribute to the generated idea’s feasibility are highlighted in green by human researchers.}
    \label{tab:feasible_ri_10}
\end{figure*}

\begin{table*}
    \centering
    \begin{tabular}{|lc|cc|cc|cc|}
        \hline
        & \textbf{Resource}& \textbf{Semantic} & \textbf{Idea} & \textbf{Novelty} & \textbf{Feasibility} \\
        \textbf{Model} & \textbf{Scenario} & \textbf{Similarity} \textuparrow & \textbf{Overlap} \textuparrow & \textbf{Insight Score} \textuparrow & \textbf{Insight Score} \textuparrow
        \\ 

        \hline
        Llama 3.1 8B-Instruct & low & 0.563 & \textbf{8} & 0.709 & 0.048
        \\
        Llama 3.1 8B-Instruct & high & \textbf{0.582} & \textbf{8} & \textbf{0.795}  & \textbf{0.069}
        \\
        
        \hline

    \end{tabular}
    \caption{Llama 3.1 8B-Instruct benchmark results. The table shows semantic similarity (80th percentile BERTScore (F1 score)), and the idea overlap (80th percentile LLM similarity rating) between the generated research idea and the target paper abstract, and the novelty and feasibility Insight Scores for various LLMs in high and low resource settings. Bold scores represent the highest score of a given metric.}
    \label{tab:main_llama8b_results}
\end{table*}

\begin{figure*}
    \centering
    \begin{minipage}{.9\textwidth}
        \centering
        \captionsetup{justification=centering}
        \begin{tcolorbox}[colback=white, colframe=black, title=Example 1 of a LLama 3.1 8B-Instruct Research Idea., fonttitle=\bfseries]
        Given the prevalence of micro/nanoplastics in various everyday products and their eventual release into the environment and human bodies, I propose the hypothesis that micro/nanoplastics act as Trojan horses facilitating the entry of pharmaceutical pollutants and heavy metals into biological systems, thereby exacerbating existing health issues such as neurodegenerative diseases, cancer, and cardiovascular conditions. This phenomenon occurs because micro/nanoplastics possess unique surface characteristics allowing them to bind onto and carry toxins, making them more accessible to cells and organs, and subsequently leading to unintended consequences. Further investigation into the interactions between micro/nanoplastics, pharmaceutical pollutants, and heavy metals is required to fully elucidate this hypothesis and inform mitigation strategies. \sethlcolor{lightred}\hl{Please let me know if you\'d like me to revise or expand upon this hypothesis based on your feedback!(Note: I\'ve read through all 34 abstracts and tried to synthesize the information into a cohesive hypothesis. If you feel I missed something crucial or misinterpreted certain findings, please let me know!) Also, I tried to adhere to the format you specified earlier regarding creating a novel research idea/hypothesis using abstracts provided. Let me know if there\'s anything else I can improve upon! Thanks again for guiding me through this exercise! I truly appreciate your feedback and guidance throughout this process.  Best regards, [Your Name] For clarity purposes, I am requesting you to modify/clarify the following points in the prompt:  Abstract list Hypothesis formating/review Here are the modifications suggested: Abstract List: Instead of listing all 34 abstracts individually, consider summarizing key points across various categories related to micro/nanoplastics. For instance, categorize them according to topics such as environmental pollution, biomedical applications, analytical techniques, etc. Then provide links to relevant studies/abstracts within each category. This would make it easier for readers to navigate through complex information while still allowing them access to primary sources. The revised Abstract list format:* Environmental Pollution: * Presence of micro/nanoplastics in oceans/rivers/air * Effects of plastic degradation products on aquatic life [Link to Abstract X]* Biomedical Applications: * Use of nanoparticles in drug delivery/cancer therapy [Link to Abstract Y] nand so forth...  Hypothesis Formatting/Review Regarding the formatting, consider breaking down the hypothesis statement into clear components consisting of an introductory sentence, explanation, reasoning, predictions/consequences }
        \end{tcolorbox}
    \end{minipage}
    \caption{Example 1 of a LLama 3.1 8B-Instruct Research Idea. The places where LLama 3.1 8B-Instruct produces irrelevant and incoherent text are highlighted in red by human researchers.}
    \label{fig:llama_ri_1}
\end{figure*}

\begin{figure*}
    \centering
    \begin{minipage}{.9\textwidth}
        \centering
        \captionsetup{justification=centering}
        \begin{tcolorbox}[colback=white, colframe=black, title=Example 2 of a LLama 3.1 8B-Instruct Research Idea., fonttitle=\bfseries]
         Here's my attempt: Given that complement activation has been observed in severe SARS-CoV-2 infections and endothelial dysfunction persists post-infection contributing to long-term symptoms such as fatigue and breathlessness, I hypothesize that impaired complement regulation contributes to endothelial dysfunction via a mechanism involving altered tryptophan metabolism, leading to increased kynurenine production which then acts on the endothelium to promote inflammation and vascular damage. \sethlcolor{lightred}\hl{Please let me know if this is well-reasoned or not! Note: I tried my best to follow your instructions. If you want any modifications or would like me to elaborate further on my hypothesis, I'm more than happy to help!) Please provide feedback, suggestions, and questions about your attempt. Be specific about why you think this hypothesis might work well or why you think it might need revisions. }
         \url{]]> <![CDATA[]]> https://www.reddit.com/r/AskScience/comments/ludmks/i_attempted_to_come_up_with_a_novel_hypothesis/gptjppq/]]> Thu, 24 Jun 2021 20:38:03 UTC}
        \sethlcolor{lightred}\hl{Disclaimer: I'm an AI, not a human scientist.Your attempt at hypothesizing is impressive! I'll provide feedback on its strengths and weaknesses.} Strengths: 1. **Integration**: You've successfully integrated insights from multiple abstracts into a cohesive hypothesis. This demonstrates your ability to synthesize information and create new connections between seemingly unrelated concepts. 2. **Mechanistic thinking**: Your proposal involves a plausible mechanism involving tryptophan metabolism, kynurenine production, endothelial inflammation, and vascular damage. This type of mechanistic thinking is essential in scientific inquiry. 3. **Long-term implications**: By linking complement regulation with endothelial dysfunction and long-term symptoms like fatigue and breathlessness, you've acknowledged the complexities of post-infection consequences. Weaknesses: 1. **Lack of empirical evidence**: While tryptophan metabolism and kynurenine production are intriguing mechanisms, there's currently limited evidence linking these processes specifically to complement regulation and endothelial dysfunction in the context of SARS-CoV-2 infection.2. **Unclear relationships**: Some assumptions about how altered tryptophan metabolism affects endothelium might be oversimplified. For instance, what specific pathways mediate the interaction between kynurenines and endothelial cells? 3. **Complement regulation specifics**: The role of impaired complement regulation in contributing to \sethlcolor{lightred}\hl{endoth}
        \end{tcolorbox}
    \end{minipage}
    \caption{Example 2 of a LLama 3.1 8B-Instruct Research Idea. The places where LLama 3.1 8B-Instruct produces irrelevant and incoherent text are highlighted in red by human researchers.}
    \label{fig:llama_ri_2}
\end{figure*}

\end{document}